\documentclass[10pt,journal,compsoc]{IEEEtran}

\ifCLASSOPTIONcompsoc
  \usepackage[nocompress]{cite}
\else
  \usepackage{cite}
\fi
\ifCLASSINFOpdf
\else
\fi

\usepackage{graphicx}
\usepackage{float}
\usepackage{subfig}
\usepackage{amsmath}
\usepackage{multirow}
\usepackage[section]{placeins}
\usepackage{float}
\usepackage{ragged2e}
\usepackage{stfloats}
\usepackage{color}
\usepackage{hyperref}
\begin{document}
%
\title{Towards Complex Backgrounds: A Unified Difference-Aware Decoder for Binary Segmentation}
%
%
%
%

\author{
Jiepan Li$^{\sharp}$,
Wei He$^{\sharp}$,~\IEEEmembership{Senior Member, IEEE,}
and~Hongyan Zhang$^{\star}$,~\IEEEmembership{Senior Member, IEEE}
\IEEEcompsocitemizethanks{\IEEEcompsocthanksitem J. Li, W. He and H. Zhang are with the State Key Laboratory of Information Engineering in Surveying, Mapping and Remote Sensing, Wuhan University, Wuhan 430072, China (\{jiepanli,weihe1990,zhanghongyan\}@whu.edu.cn).
\protect\\
}
\thanks{Manuscript received XXXX; revised XXXX.}}

\markboth{Journal of \LaTeX\ Class Files,~Vol.~14, No.~8, August~2015}%
{Shell \MakeLowercase{\textit{et al.}}: Bare Advanced Demo of IEEEtran.cls for IEEE Computer Society Journals}

\IEEEtitleabstractindextext{%
\begin{abstract}
\justifying
Binary segmentation is used to distinguish objects of interest from background, and is an active area of convolutional encoder-decoder network research. The current decoders are designed for specific objects based on the common backbones as the encoders, but cannot deal with complex backgrounds. Inspired by the way human eyes detect objects of interest, a new unified dual-branch decoder paradigm named the difference-aware decoder is proposed in this paper to explore the difference between the foreground and the background and separate the objects of interest in optical images. The difference-aware decoder imitates the human eye in three stages using the multi-level features output by the encoder. In Stage A, the first branch decoder of the difference-aware decoder is used to obtain a guide map. The highest-level features are enhanced with a novel field expansion module and a dual residual attention module, and are combined with the lowest-level features to obtain the guide map. In Stage B, the other branch decoder adopts a middle feature fusion module to make trade-offs between textural details and semantic information and generate background-aware features. In Stage C, the proposed difference-aware extractor, consisting of a difference guidance model and a difference enhancement module, fuses the guide map from Stage A and the background-aware features from Stage B, to enlarge the differences between the foreground and the background and output a final detection result. To verify the performance of the proposed difference-aware decoder, we choose four well known backbones including VGG, ResNet, Res2Net, PVT, and four binary segmentation tasks, $i.e.$, salient object detection, camouflaged object detection, polyp segmentation, and mirror detection, for comparative experiments. The results demonstrate that the difference-aware decoder can achieve a higher accuracy than the other state-of-the-art binary segmentation methods for these tasks. The source code will be available on \href{https://github.com/Henryjiepanli/DAD}{https://github.com/Henryjiepanli/DAD}.
\end{abstract}

}

\maketitle

\IEEEdisplaynontitleabstractindextext

\IEEEpeerreviewmaketitle

\ifCLASSOPTIONcompsoc
\IEEEraisesectionheading{\section{Introduction}\label{sec:introduction}}
\else
\section{Introduction}
\label{sec:introduction}
\fi
\IEEEPARstart{B}{inary} segmentation is aimed at distinguishing pixels belonging to the foreground from the background, which is a problem that has attracted much attention in the field of computer vision. Binary segmentation has been extended to various tasks, such as salient object detection (SOD)\cite{zhao2019egnet}, camouflaged object detection (COD)\cite{fan2020camouflaged}, polyp segmentation\cite{fan2020pranet}, and mirror detection\cite{yang2019my}. It has also been applied in autonomous driving\cite{mahadevan2009saliency}, robot navigation\cite{craye2016environment}, disaster assessment\cite{valentijn2020multi}, medical diagnosis ($e.g.$, lung infection segmentation \cite{fan2020inf}, polyp segmentation \cite{fan2020pranet}), and even the military applications\cite{lin2019metaheuristic}.

\par
\begin{figure}[t]
\vspace{-1.0em}
\centering
\subfloat[Image]{
\includegraphics[width=2.1cm,height=5.5cm]{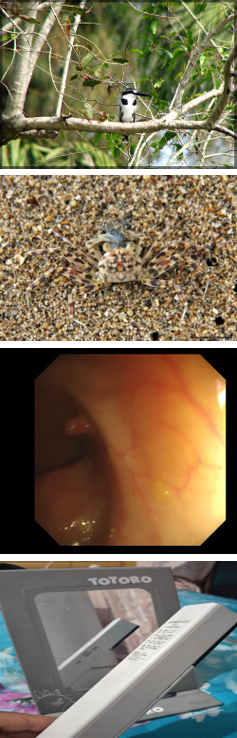}%
\label{fig_vs_image}}
\hfil
\subfloat[SOTA]{
\includegraphics[width=2.1cm,height=5.5cm]{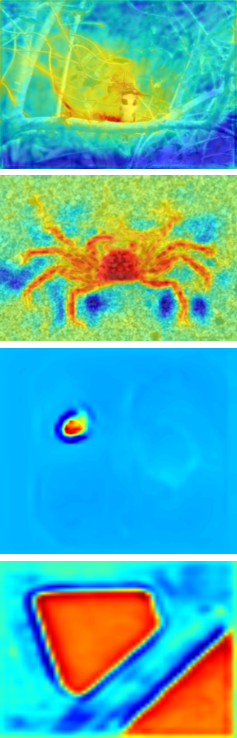}%
\label{fig_vs_sota}}
\hfil
\subfloat[DAD]{
\includegraphics[width=2.1cm,height=5.5cm]{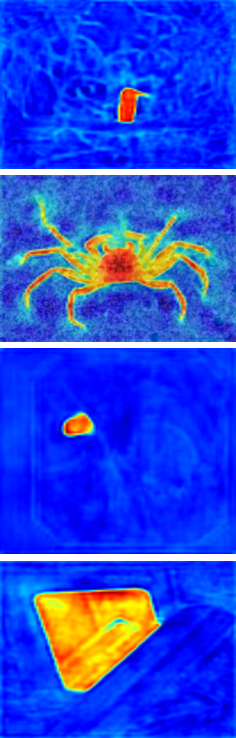}%
\label{fig_vs_ours}}
\hfil
\subfloat[Groundtruth]{
\includegraphics[width=2.1cm,height=5.5cm]{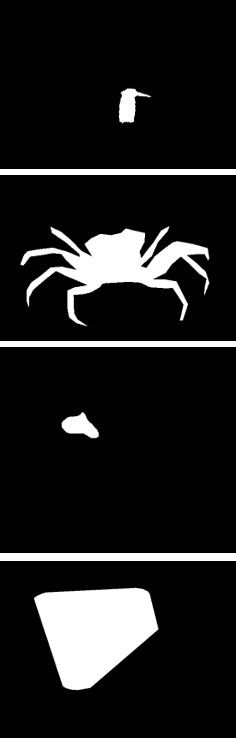}%
\label{fig_vs_gt}}
\hfil
\caption{Visual examples from the test dataset. Experiments were conducted in salient object detection (SOD) (1st row), camouflaged object detection (COD) (2nd row), polyp segmentation (3rd row), and mirror detection (4th row). Heatmaps of the difference-aware decoder (DAD) and the corresponding state-of-the-art (SOTA) methods are visualized (BASNet\cite{DBLP:journals/corr/abs-2101-04704}, SINet V2\cite{9444794}, Polyp-PVT\cite{dong2021polyp}, and MirrorNet\cite{yang2019my}).}
\vspace{-2em}
\label{fig_1}
\end{figure}
In these tasks and applications, the foreground objects can be either similar to or very different from the background, making binary segmentation a difficult task. This task can be solved by the use of a convolutional neural network (CNN)-based model\cite{krizhevsky2012imagenet}, as these models have outstanding feature representation capabilities and show a strong generalization performance. In these networks, the encoder-decoder structure\cite{long2015fully} has become a major paradigm, where the encoder extracts multi-level features by performing a series of convolution operations on an input image, and the decoder\cite{Zhang_2018_ECCV} takes responsibility for utilizing the multi-level features to generate the segmentation results. Typically, the commonly used encoders (which are also called backbones) include VGG\cite{simonyan2014very}, ResNet\cite{he2016deep}, MobileNet\cite{howard2017mobilenets}, Res2Net\cite{gao2019Res2Net}, and the recently proposed pyramid vision transformer (PVT)\cite{wang2021pyramid}. The low-level features \cite{wu2019cascaded} are rich in detailed information (texture, color, etc.), but their semantic information is insufficient. Meanwhile, the high-level features \cite{wu2019cascaded} are highly semantic, but are lacking in detailed information, due to the downsampling. Therefore, the key to binary segmentation is how to take advantage of these multi-level features in the decoder design, which is a problem that has been well studied over the past few years.

\par

The general decoding strategy of binary segmentation is to aggregate multi-level encoded features from the aspects of expanding the receptive field (\cite{chen2017deeplab,yu2015multi,chen2018encoder}), enriching the contextual information (\cite{zhang2018context,zhao2017pyramid,peng2017large,he2019adaptive}), or feature refinement (\cite{fu2019dual,lin2017refinenet}), to improve the final segmentation performance. Specifically, for the SOD task (\cite{wu2019cascaded,hou2017deeply,liu2018picanet,yang2019my}), the decoder is designed to highlight the salient features from the decoding process, whereas, for the COD task (\cite{mei2021camouflaged,9008371,li2021uncertainty,lv2021simultaneously}), the decoder is applied to simulate animal hunting in nature with the procedure of searching for prey and then capturing the prey. However, as shown in Fig. \ref{fig_1}, the complex backgrounds of the various binary segmentation tasks increase the difficulty of efficient feature extraction. In summary, although various decoding strategies have been specifically developed to highlight the object features from the encoder for different binary segmentation tasks, how to design a general efficient decoder considering various backgrounds still remains a challenge.


\par
In  order to deal with the various complex backgrounds, we propose a unified \textbf{dual-branch decoder paradigm named the difference-aware decoder (DAD) for the binary segmentation task}. This is inspired by the procedure of how human eyes detect special objects \cite{lin2014evaluating}. Specifically, the eyes first observe the foreground objects and obtain a coarse detection map of the objects of interest. Secondly, the eyes pay substantial attention to the background area adjacent to the foreground objects. Finally, the differences between the foreground and the background are enlarged by the processing of the brain, and the real objects can be precisely outlined. To imitate the procedure of human eyes, on the basis of the multi-level features from the encoder, the proposed difference-aware decoder has the following three stages. \textbf{In Stage A, the first branch decoder} adopts the general decoding strategies to \textbf{obtain a coarse guide map}. In detail, the highest-level features are enhanced with a novel field expansion module (FEM) and a dual residual attention (DRA) module\cite{fu2019dual}, and are then combined with the lowest-level features to obtain the coarse guide map. \textbf{In Stage B, the other branch decoder} adopts a middle feature fusion (MFF) module in order to trade-off the textural details, take the semantic information into account, and \textbf{obtain the background-aware features}. \textbf{In Stage C, the proposed difference-aware extractor (DAE)}, which consists of a difference guidance model (DGM) and a difference enhancement module (DEM), fuses the foreground guide map from Stage A and the background-aware features from Stage B, to \textbf{enlarge the differences between foreground and background and output the final segmentation results}. The proposed difference-aware decoder method benefits from the differences between the guide map and the background-aware features, and \textbf{can overcome the variable background} while achieving a superior performance, as illustrated in Fig. \ref{fig_1}.
\par
The rest of this paper is organized as follows. Section 2 introduces the related work. Section 3 presents the full details of the proposed difference-aware decoder method. The comprehensive experimental results and the analysis, including a parametric analysis and ablation study, are provided in Section 4. Section 5 draws the conclusions from this study. Due to the page limitation, further details about the experiments are presented in the Supplementary Material.
\vspace{-0.3em}
\section{RELATED WORKS}
In this paper, we mainly review the development of deep learning methods for two typical binary segmentation tasks, $i.e.$, SOD and COD. We also review the development of the dual-branch decoder for various applications. For more binary segmentation tasks, we refer the reader to \cite{fan2020pranet},\cite{huang2021hardnet},\cite{zhang2020adaptive},\cite{patel2021enhanced},\cite{wei2021shallow},\cite{dong2021polyp}, and \cite{yang2019my}.

\subsection{Salient Object Detection}

CNNs can be used to locate the boundary of detected salient regions and conduct the segmentation. For example, Zhao \textit{et al.}\cite{zhao2015saliency} utilized a CNN to simulate the saliency in the image by modeling the global context, and modeled the local context for the saliency prediction in refined regions; Li \textit{et al.} \cite{li2015visual} combined the multi-level features extracted by a CNN with a network with multiple fully connected layers, which can achieve a very high level of regression accuracy; and Liu \textit{et al.}\cite{liu2016dhsnet} utilized an end-to-end network named DHSNet to realize hierarchical detection of significant targets from global to local and coarse to fine. Other researchers have focused on the deficiencies in the precise localization of high-level features and have proposed some improved approaches\cite{lee2016deep}. However, the CNN-based methods have some problems, such as blurred and inaccurate predictions near the boundaries of salient objects\cite{borji2019salient}.

\par
\begin{figure*}[htb]
\centering
\includegraphics[width=1.0\linewidth]{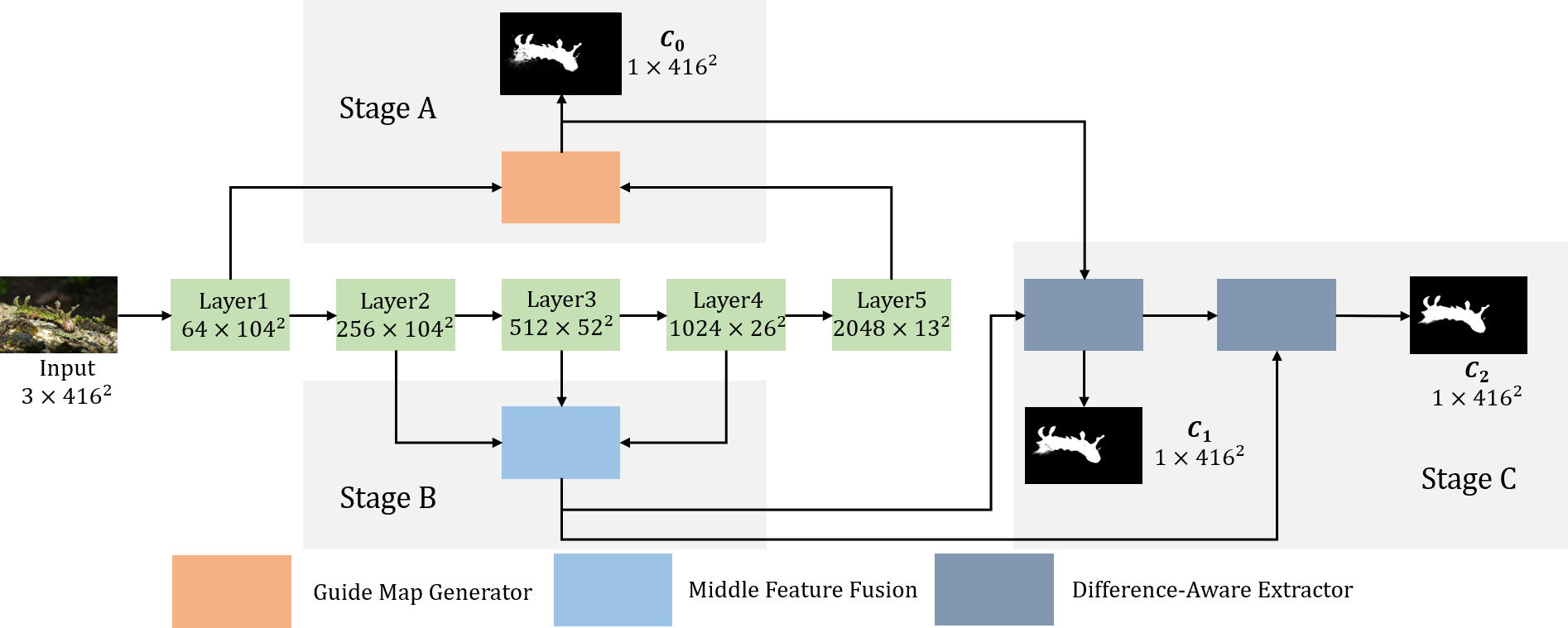}
\caption{The structure of the difference-aware decoder. On the basis of the multi-level features extracted by the common backbones, such as VGG, ResNet, Res2Net, and PVT, the whole process is divided into three stages: the guide map generator (GMG), the middle feature fusion (MFF) module, and the difference-aware extractor (DAE). The GMG attempts to obtain a coarse guide map, the MFF module attempts to find the background-aware features, and the DAE attempts to fuse the guide map and the background-aware features, to enhance the difference between the foreground and the background, and obtain a refined detection result.}
\label{fig_2}
\vspace{-1.0em}
\hfil
\end{figure*}

Subsequently, the fully convolutional network (FCN)\cite{long2015fully} with encoder-decoder structure was successfully introduced for the SOD task. Based on the common encoders, researchers have developed various decoders to interpret the multi-level features. For example, Wu \textit{et al.}\cite{wu2019cascaded} put forward a cascaded partial decoder to suppress the distractors in the features and improve the feature representation ability; Liu \textit{et al.}\cite{liu2018picanet} decoded the high-level features with the low-level features by proposing a feature aggregation model; Zhao \textit{et al.}\cite{zhao2019egnet} designed a decoder to integrate the local edge information and global location information to obtain the salient edge features;  Qin \textit{et al.}\cite{qin2019basnet} proposed a densely supervised encoder-decoder network and utilized a residual refinement module to refine the final saliency map; Pang \textit{et al.}\cite{pang2020multi} proposed a decoder embedded with self-interaction modules to obtain more efficient multi-scale features from the integrated features; Wang \textit{et al.}\cite{wang2019salient} used a pyramid attention structure to concentrate more on salient regions while considering multi-scale saliency information; and Zhao \textit{et al.}\cite{zhao2020suppress} designed a decoder based on the consideration of the disparity of the contributions of the different encoder blocks. However, in general, \textbf{most of the above methods can distinguish the objects of interest from a simple background, but they are incapable of discovering the objects of interest in the case of a complex background}, especially for camouflaged objects. In this paper, we investigate how best to deal with complex backgrounds for various binary segmentation tasks.

\subsection{Camouflaged Object Detection}

The COD task is much more difficult than the SOD task. The development of encoder-decoder based COD methods can be classified into two aspects, $i.e.$, encoders and decoders. With regard to encoders, ResNet\cite{fan2020camouflaged} was the first to be introduced, which was followed by Res2Net\cite{9444794} and PVT\cite{liu2022boosting}. With regard to decoders, SINet\cite{fan2020camouflaged} simulates the predation process in nature and searches for and identifies the potential objects of interest. PFNet\cite{Mei_2021_CVPR} improves the search for potential objects with a global positioning module and the identification process with a focus module. In addition,  Lv \textit{et al.} \cite{lv2021simultaneously} suggested that explicitly modeling the estimation of the conspicuousness of a camouflaged object against its surrounding can not only better explain animal camouflage and evolution, but can also provide guidance for designing more complex camouflage techniques. On this basis, Lv \textit{et al.} \cite{lv2021simultaneously} proposed a ranking-based COD decoder. Zhai \textit{et al.}\cite{zhai2021mutual} designed a mutual graph learning (MGL) decoder model to generalize the idea of conventional mutual learning from regular grids to the graph domain. Liu \textit{et al.} \cite{sun2021context} integrated the multi-level features with informative attention coefficients and obtained multi-scale feature representations for exploiting the rich global context information. On the basis of transformer backbones, more advanced decoders have also been developed \cite{liu2022boosting}. Although simulating the predation process is a relatively efficient approach for COD, it \textbf{mainly focuses on the foreground camouflaged objects and ignores the contributions from the background}. The proposed difference-aware decoder realizes the importance of the difference between the foreground and background, to imitate the object detection procedure of human eyes.

\subsection{Dual-branch Decoder Architectures}

The dual-branch decoder architecture has been introduced into various binary segmentation tasks, to obtain a better accuracy. For example, Wu \textit{et al.}\cite{wu2019cascaded} designed a dual-branch decoder named the cascaded partial decoder, which utilizes the initial saliency map of the first branch to refine the features of the second branch. In addition, Fan \textit{et al.}\cite{fan2020camouflaged,9444794} proposed a dual-branch decoder that uses a holistic attention mechanism to refine the segmentation map in the first decoder and output a more accurate result. Other works have utilized dual decoders for complementary tasks to improve the learning efficiency and generalization across different tasks. For example, Zhang \textit{et al.}\cite{zhang2019pattern} utilized a joint task-recursive learning framework to refine the results of both semantic segmentation and monocular depth estimation through serialized task-level interactions; Zhen \textit{et al.}\cite{zhen2020joint} proposed a dual-branch decoder framework to fuse the feature maps generated for semantic segmentation and boundary detection; and Li \textit{et al.} \cite{li2021uncertainty} used a dual-branch decoder to achieve both a higher-order similarity measure and network confidence estimation. Typically, the dual-branch decoder acts as different approaches to achieve the segmentation task.

\par
 In this paper, \textbf{we also propose a unified dual-branch decoder architecture to imitate the object detection procedure of human eyes for binary segmentation}. Specifically, one branch is used to discover the coarse foreground, while the other is used to exploit the background-aware features. Finally, the guide map and the background-aware features are fused to extract the difference-aware features for the final binary object segmentation.

\section{METHODOLOGY}
\subsection{Overview}
In this section, we describe the proposed difference-aware decoder in detail, as illustrated in Fig. \ref{fig_2}. The process of how objects are detected by human eyes is imitated on the basis of the multi-level features from the backbone. Three stages are utilized to simulate the procedure\cite{lin2014evaluating} of how human eyes detect special objects. To be specific, Stage A outputs a guide map, which can represent the coarse foreground information; Stage B fuses the three middle-level features to obtain the background-aware features; and Stage C makes use of the guide map and the background-aware features to enhance the difference between the foreground and the background and obtain a refined detection map.

\begin{figure}[!t]
\centering
\includegraphics[width=3.5in]{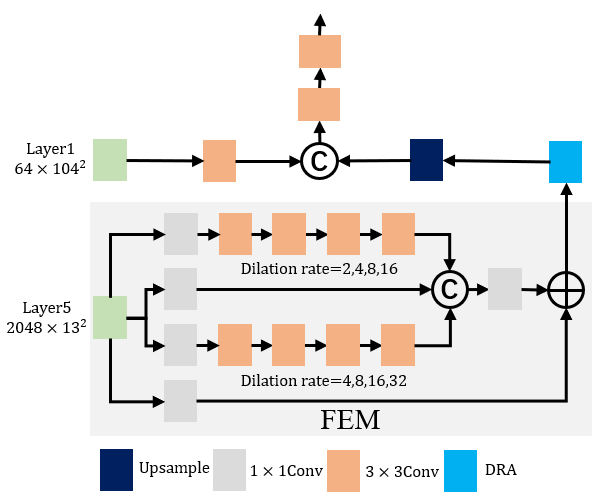}
\caption{The architecture of the guide map generator (GMG).}
\label{fig_gmg}
\vspace{-1.0em}
\end{figure}
\subsection{Stage A: Guide Map Generator}
As presented in Fig. \ref{fig_2}, in Stage A, the aim is to obtain a guide map of the objects of interest, which is achieved by the GMG in the difference-aware decoder. Fig. \ref{fig_gmg} illustrates the details of the proposed GMG architecture, with the highest- and lowest-layer encoded features as input. It has been reported that CNNs are apt to extracting information mainly from the much smaller regions in the receptive field\cite{zhou2014object}, since the receptive field helps to highlight the importance of the regions closer to the center and elevates the insensitivity to small spatial shifts. However, in the binary segmentation task, even the increase of the encoding layers can enlarge the receptive field, which is an inefficient way to cover the camouflaged/salient objects in the task, limiting the final performance. Therein, we propose the GMG to enlarge the receptive field as much as possible, to cover the foreground objects by atrous convolution\cite{chen2014semantic}, while keeping the feature map resolution unchanged. To further enrich the detailed information of the objects, the lowest-layer encoding features are combined with the enlarged features to generate the guide map.
\par
As shown in Fig. \ref{fig_gmg}, on the basis of the multi-scale features extracted from the backbone, the FEM consists of four parallel paths to enlarge the receptive field. The first path is made up of a $1\times1$ convolution block and four consecutive $3\times3$ convolution blocks. Importantly, the dilation rates of those four $3\times3$ convolution blocks are 4, 8, 16, and 32, respectively. The second path is the simple $1\times1$ convolution block to reduce the dimensions of the input features. The third path is similar to the first branch, with the dilation rates of the $3\times3$ convolution blocks replaced with 2, 4, 8, and 16 in turn. The outputs of the first three paths are then concatenated to fuse the features with sufficient contextual information, followed with a $1\times1$ convolution block. Finally, a modified residual structure is adopted to enhance the final output of the FEM, which is implemented via a $1\times1$ convolution block to match the number of concatenated features. It is worth emphasizing that all the convolution blocks are composed of a convolutional layer, a batch normalization layer, and an activation layer.
\begin{figure}[!t]
\centering
\subfloat[ASPP]{
\includegraphics[width=2.1cm,height=2.1cm]{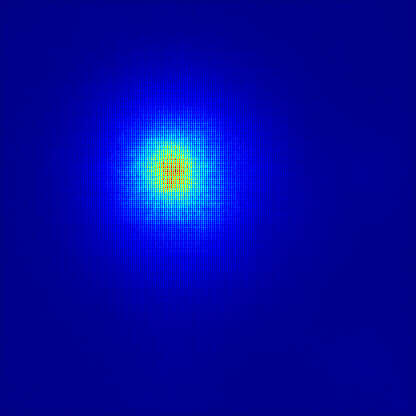}%
\label{fig_ASPP}}
\hfil
\subfloat[RFB]{
\includegraphics[width=2.1cm,height=2.1cm]{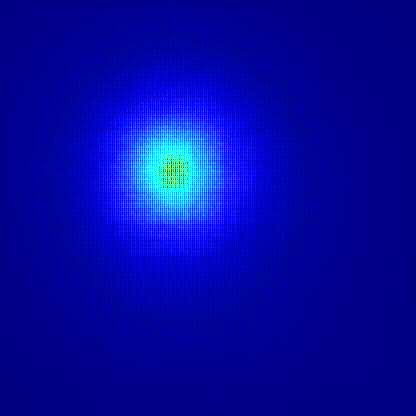}%
\label{fig_RFB}}
\hfil
\subfloat[DenseASPP]{
\includegraphics[width=2.1cm,height=2.1cm]{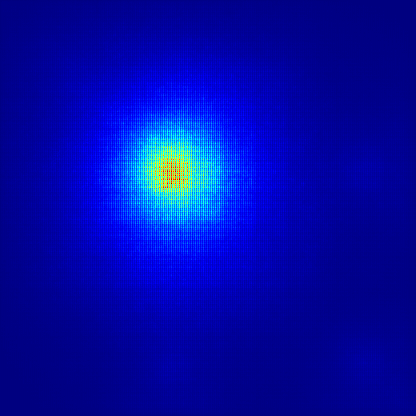}%
\label{fig_Denseaspp}}
\hfil
\subfloat[FEM]{
\includegraphics[width=2.1cm,height=2.1cm]{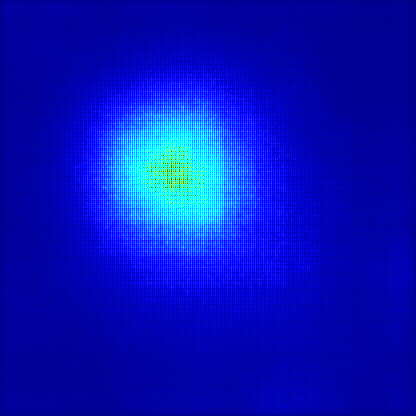}%
\label{fig_FEM}}
\hfil
\caption{Comparison of the FEM with three typical modules (atrous spatial pyramid pooling (ASPP)\cite{chen2017deeplab}, the receptive field block (RFB) \cite{liu2018receptive}, and DenseASPP\cite{yang2018denseaspp}). The FEM was replaced with these three typical modules, and the visualization of each module is the effective receptive field derived from the correspondent layer in the trained model, depicted by the gradient back-propagation method in \cite{luo2016understanding}.}
\label{fig_fem_compare}
\vspace{-1.0em}
\end{figure}
\par
Previous works, such as ASPP\cite{chen2017deeplab}, RFB\cite{liu2018receptive}, and DenseASPP\cite{yang2018denseaspp}, have proved the importance of enlarging the receptive field. Differently, in the proposed approach, the dilation rate is increased in the consecutive $3\times3$ convolution blocks to increase the receptive field of the path as much as possible. What is more, the two paths with simple $1\times1$ convolution blocks can keep the original information from both the channel dimension and the spatial dimension. As shown in Fig. \ref{fig_fem_compare}, the FEM represents an effective way to make use of this parallel structure, to obtain a more appropriate receptive field.
\par
Aiming at modeling the long-range dependencies to capture rich contextual relationships for better feature representations with intra-class compactness, the attention module is further utilized to enhance the output FEM features with an enlarged receptive field\cite{fu2019dual}. In detail, the DRA module from\cite{fu2019dual} is introduced to enhance the highest-level features. By this sequential operation of the FEM and DRA\cite{fu2019dual}, the global information can be captured and the long-distance dependency between the foreground and the background can be established. This reduces the interference from extraneous information, which is of huge benefit to binary segmentation, especially in some difficult tasks. Subsequently, the enhanced features obtained via the DRA module are combined with the lowest-layer encoded features to provide enough detailed information. The highest-level features are upsampled to the same size as the lowest-level features. These two types of features are then concatenated and processed with two $3\times3$ convolution blocks, to finally obtain the guide map $C_{0}$, which will be further used for the guided loss calculation.

\subsection{Stage B: Middle Feature Fusion}
As presented in Fig. \ref{fig_2}, Stage B involves extracting the background-aware features by the proposed MFF strategy, with the three middle-level features as inputs. As mentioned previously, in the object detection procedure of human eyes, the background information is important to outline the foreground objects of interest. However, for the binary segmentation task, the background extraction is a challenge due to the complexity of the background compared to the foreground. From another aspect, the lowest-level encoding features are more detailed and result in a complex background, whereas the highest-level encoding features are semantic and lose the detailed background information. Therein, we chose the three middle-level encoding features for the background-aware feature extraction.
\par
\begin{figure}[!t]
\centering
\includegraphics[width=3.5in]{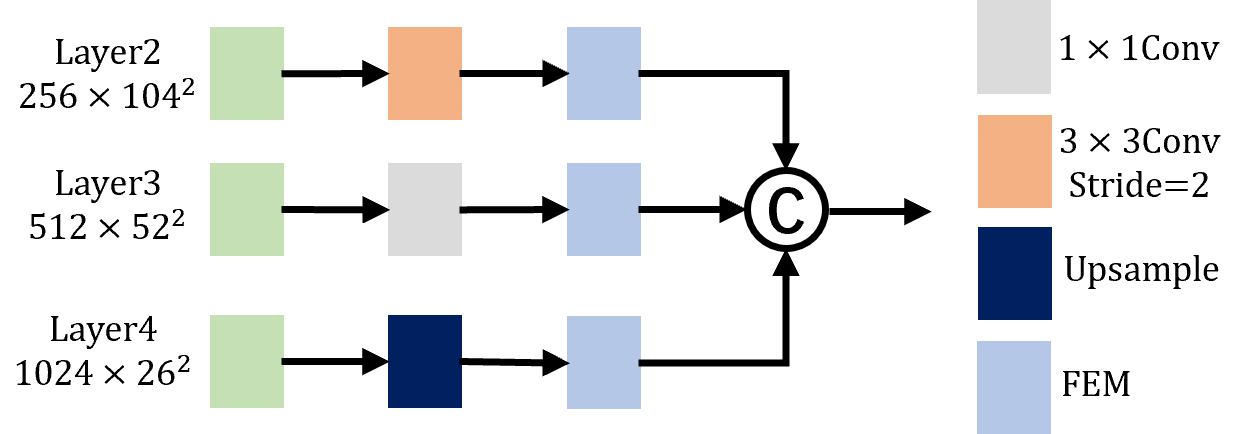}
\caption{The pipeline of the middle feature fusion (MFF) module. The layers are obtained from the backbone.}
\label{fig_MFF}
\vspace{-1.0em}
\end{figure}

Differing from the mainstream architecture of the top-down and bottom-up methods, which resize the multi-level features to the biggest size or the smallest size, we propose fusing the three features of the middle-level size. As shown in Fig. \ref{fig_MFF}, the MFF module selects the middle size as the base size and resizes the other features to fit this base size. After using an upsampling operation for layer 4, this is resized to fit the size of layer 3. One $3\times3$ convolution block and two $1\times1$ convolution blocks are utilized to reduce the channels of layer 2, layer 3, and layer 4 to the same number (which is set to 32). The stride of the $3\times3$ convolution block is 2, so it can act as a downsampling operation. Before concatenating these three processed layers, the FEM is used to obtain a larger receptive field with richer contextual information. Finally, the concatenated three-layer features are utilized to represent the background-aware features.

\par
\subsection{Stage C: Difference-Aware Extractor}
\begin{figure}[!t]
\centering
\includegraphics[width=3.5in]{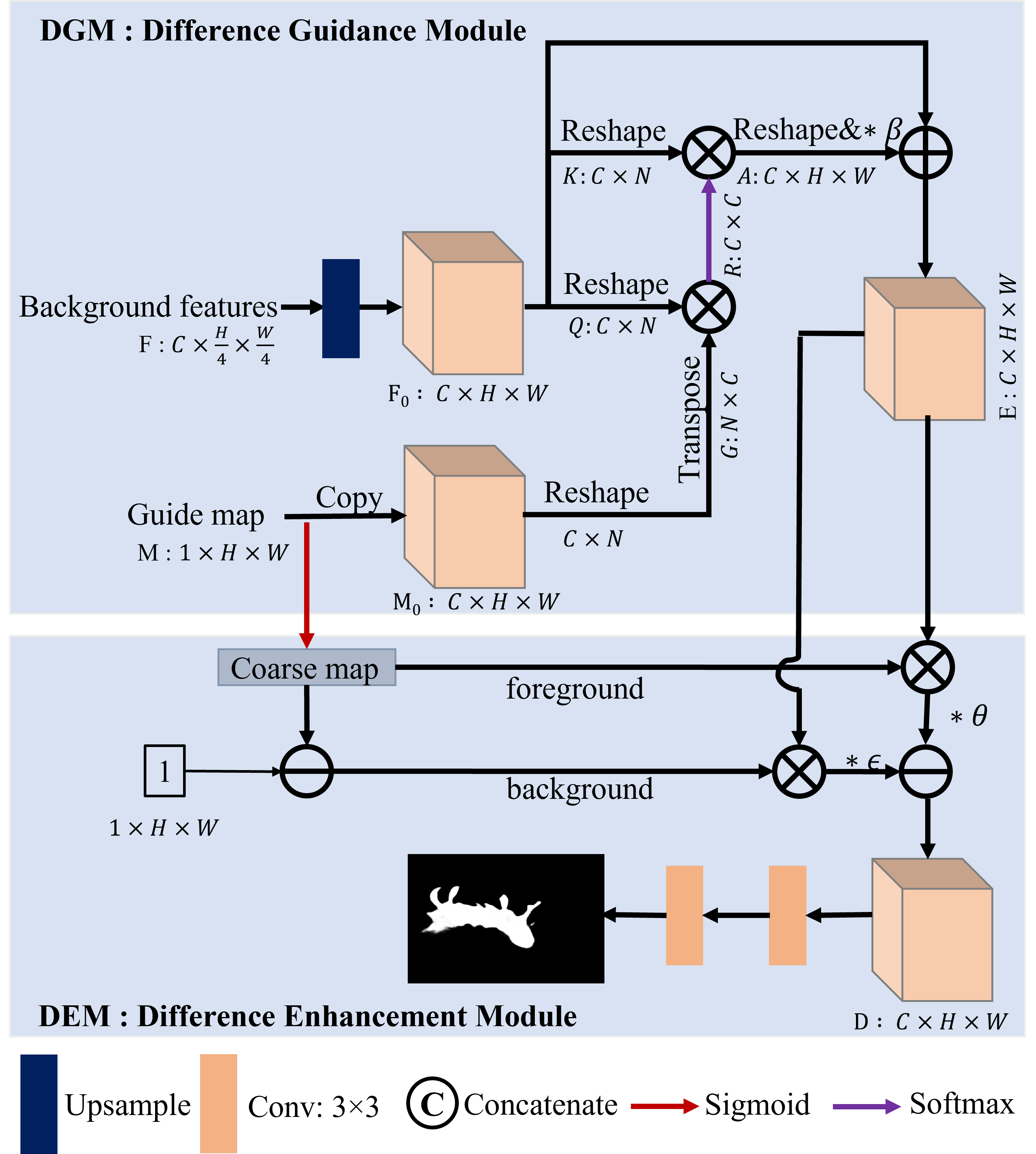}
\caption{The pipeline of the difference-aware extractor (DAE) module, which is composed of a difference guidance module (DGM) and a difference enhancement module (DEM).}
\label{fig_DAE}
\vspace{-1.0em}
\end{figure}

In the first two stages, the two-branch decoder is used to obtain the guide map and the background features. Therefore, in Stage C, as presented in Fig. \ref{fig_2}, we are committed to exploring the differences between the foreground and the background for the final segmentation of the objects of interest, which is achieved by the proposed DAE. The DAE can take advantage of the prior knowledge of the guide map to guide the background-aware features to learn more subtle differences. As shown in Fig. \ref{fig_DAE}, the DAE consists of a DGM and a DEM. The DGM fuses the guide map and background features to separate the foreground objects from the background. The DEM then separately learns the foreground and background-aware features, and then fuses them adaptively to generate the final refined segmentation map. The details of the DGM and DEM are presented in the following.

\subsubsection{Difference Guidance Module}
As presented in Fig. \ref{fig_DAE}, the inputs of the DGM are the guide map $M \in  {R}^ {1\times H\times W}$ and the background-aware features $F \in{R}^ {C\times \frac H4\times \frac W4}$ from the two branches.
\par
Firstly, the background-aware feature $F$ is upsampled to the same sample size as $M$, while the copy operation is performed on $M$ to extend the channel number of the guide map to the same as that of $F$. The formulation is as follows:
\begin{equation}
	\label{equ1}
	\begin{split}
		F_0 &= Upsample(F), \\
		M_0 &= Copy(M).
	\end{split}
\end{equation}
\par
Secondly, the cross-attention operation is performed between $F_0$ and $M_0$. $M_0$ is reshaped and transposed to be $G \in {R}^ {N\times C}$ and $F_0$ is reshaped to obtain two feature maps $Q\in {R}^ {C\times N}$ and $K\in {R}^ {C\times N}$. After this, the guidance is performed between $Q$ and $G$. In detail, matrix multiplication is performed between $Q$ and $G$ and the result is reshaped to obtain the map $R\in {R}^ {C\times C}$. $R$ is then processed by softmax normalization to obtain the probability value, which represents the relationship of the difference between the foreground and the background. The process can be summarized as:
\begin{equation}
	\label{equ2}
	\begin{split}
		R_i{}_j &= \frac{exp(Q_i\times G_j)}{\sum_{j=1}^C exp(Q_i\times G_j)}.
	\end{split}
\end{equation}
Meanwhile, matrix multiplication is used between $K$ and $R$ and the result is reshaped to the attention features $A \in {R}^ {C\times H\times W}$. Finally, a residual structure is introduced to obtain the final enhanced  $E \in {R}^ {C\times H\times W}$:
\begin{equation}
	\label{equ3}
	\begin{split}
		E &= \beta \times A + F_0,
	\end{split}
\end{equation}
where $\beta$ is a learnable parameter.

\subsubsection{Difference Enhancement Module}
As shown in Fig. \ref{fig_DAE}, the DEM is used to separably learn the foreground and background features, while adaptively fusing the two to enhance the difference between the foreground and the background. As mentioned before, the guide map can be directly used to generate the coarse map $P \in {R}^ {1\times H\times W}$ by using the sigmoid function. In the coarse map $P$, a pixel value closer to one means that the pixel belongs to the foreground, to a large extent, and vice versa. Therefore, $P$ and $1-P$ are used to represent the probability of being foreground and background, respectively. These two probability maps are then used to extract the features from $E$:
\begin{equation}
	\label{equ4}
	\begin{split}
		D_f &= P \times E, \\
		D_b &= (1-P) \times E,
	\end{split}
\end{equation}
where $D_f \in {R}^ {C\times H\times W}$ represents the foreground features, and $D_b \in {R}^ {C\times H\times W}$represents the background features. To enlarge the difference between the foreground and the background, subtraction between $D_f$ and $D_b$ is utilized. Furthermore, two learnable parameters  $\theta$ and $\epsilon$ are introduced to adaptively fuse the two types of features, and obtain the difference-aware features $D \in {R}^ {C\times H\times W}$:
\begin{equation}
	\label{equ5}
	\begin{split}
		D &= \theta \times D_f - \epsilon \times D_b. \\
	\end{split}
\end{equation}
In fact, the subtraction used in Eq.\ref{equ5} can be replaced with addition, due to the introduction of the two learnable parameters. Finally, the difference-aware features $D$ are processed by two $3\times3$ convolution blocks and converted to the refined map by the sigmoid function.

\subsection{Loss Function}
To enhance the difference between the foreground and the background as much as possible, the DAE module is utilized twice. As shown in Fig. \ref{fig_2}, the output of the DAE is regarded as the guide map, which can again be merged with the background-aware features and processed by the DAE. The DAE is used twice, and the output of the first DAE is processed by the sigmoid function to produce the map $C_1$, while the second DAE produces the refined map  $C_2$. At this point, there are three output maps $C_0$, $C_1$, and $C_2$, which can be used to build the loss function with the ground truth. Weighted binary cross-entropy (BCE) loss and weighted intersection over union (IoU) loss are used to construct the loss function\cite{wei2020f3net}, $i.e.,$ $loss=loss_{wbce} + loss_{wiou}$ for each output map. Therefore, the total loss function can be described as:
\begin{equation}
	\label{equ6}
	\begin{split}
		loss & = \sum_{i=0}^2loss(C_i,GT).
	\end{split}
\end{equation}

\subsection{Analysis of the Three Stages}
\begin{figure}[t]
\centering
\subfloat[Image]{
\includegraphics[width=3cm]{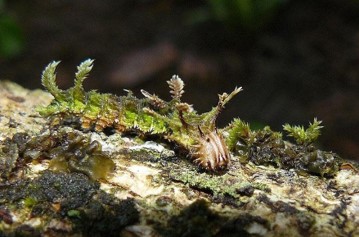}%
\label{fig_anl_1}}
\subfloat[GT]{
\includegraphics[width=3cm]{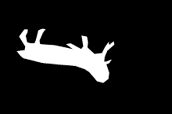}%
\label{fig_anl_2}}
\subfloat[guide map]{
\includegraphics[width=3cm]{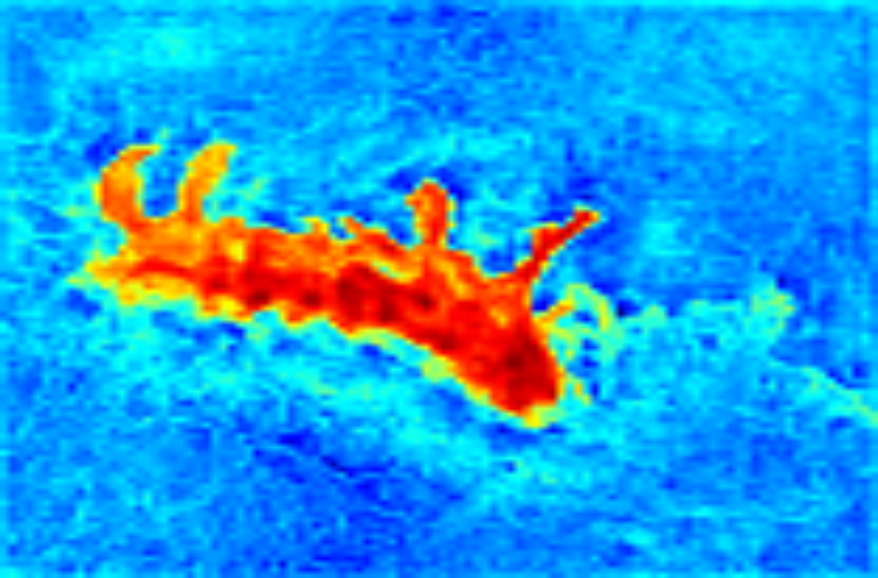}%
\label{fig_anl_3}}
\newline
\vspace{-1.0em}
\subfloat[MFF]{
\includegraphics[width=3cm]{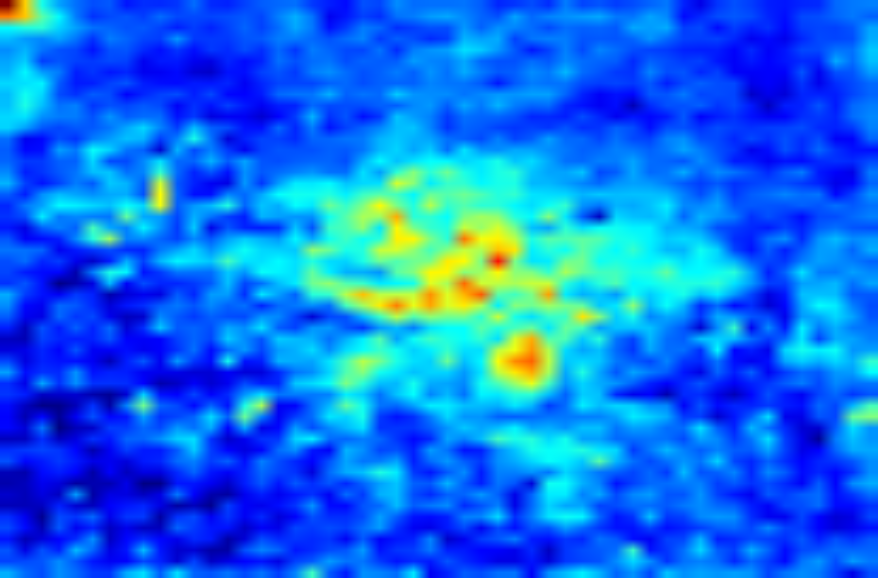}%
\label{fig_anl_4}}
\subfloat[DGM]{
\includegraphics[width=3cm]{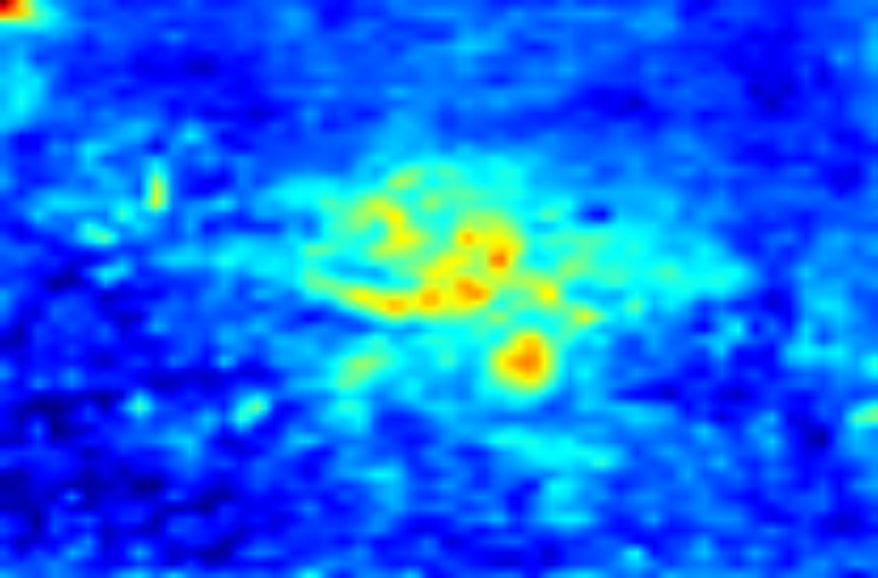}%
\label{fig_anl_5}}
\subfloat[DEM]{
\includegraphics[width=3cm]{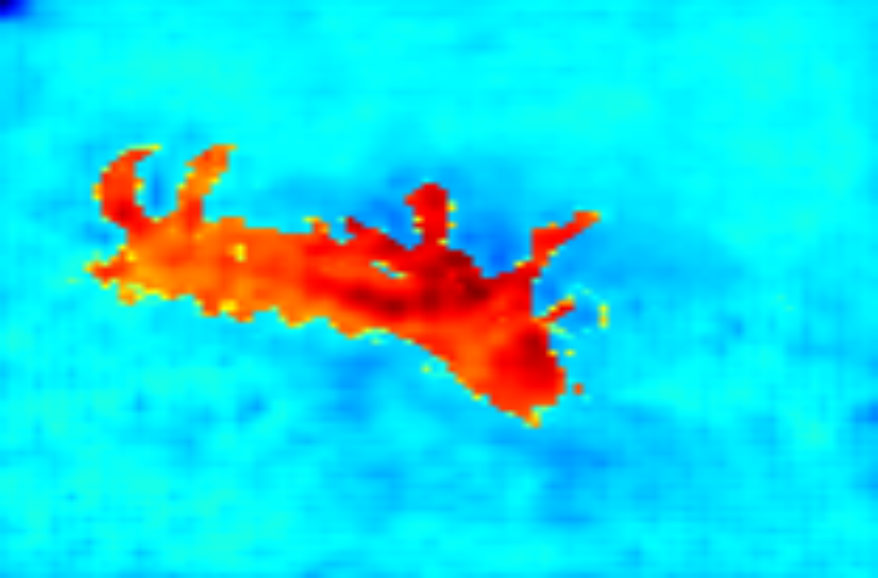}%
\label{fig_anl_6}}
\caption{The feature visualization of each stage in the proposed method.}
\label{fig_anl}
\vspace{-1.0em}
\end{figure}

To verify the proposed paradigm, the feature maps of each stage were visualized in a real-data experiment. As can be seen in Fig. \ref{fig_anl} (c) and (d), it is clear that the output of the guide map is a coarse segmentation result for the camouflaged objects, whereas the MFF module produces the background-aware features. The proposed DAE, consisting of the DGM and DEM, is then used to exploit the differences between the two. In detail, the DGM outputs the enhanced background features, as shown in Fig. \ref{fig_anl} (e), and the DEM outputs the difference-aware features, as shown in Fig. \ref{fig_anl} (f). By comparing Fig. \ref{fig_anl} (c) and (f), it can be concluded that the binary object with fused difference-aware features is much clearer than the coarse map, indicating the advantage of the proposed paradigm.

\section{Experiments}
\begin{figure*}[t]
\centering
\subfloat[Image \protect \\Backbone]{
\includegraphics[width=1.7cm,height=3cm]{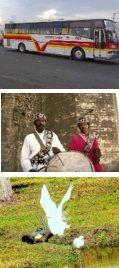}%
\label{fig_image}}
\hfil
\subfloat[GT]{
\includegraphics[width=1.7cm,height=3cm]{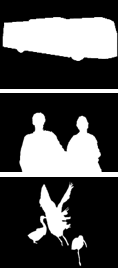}%
\label{fig_gt}}
\hfil
\subfloat [GateNet \protect\\ ResNet-50]{
\includegraphics[width=1.7cm,height=3cm]{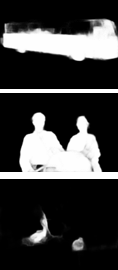}
\label{fig_gatenet}}
\hfil
\subfloat [MINet \protect\\ ResNet-50]{
\includegraphics[width=1.7cm,height=3cm]{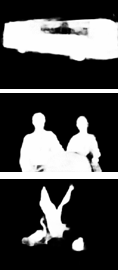}%
\label{fig_minet}}
\hfil
 \subfloat[PoolNet \protect\\ ResNet-50]{
\includegraphics[width=1.7cm,height=3cm]{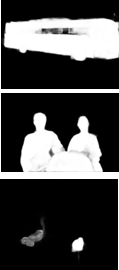}%
\label{fig_poolnet}}
\hfil
\subfloat [PoolNet + \protect\\ ResNet-50]{
\includegraphics[width=1.7cm,height=3cm]{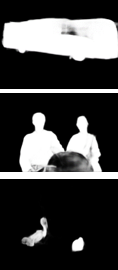}%
\label{fig_poolnet+}}
\hfil
 \subfloat[CSF \protect\\ Res2Net-50]{
\includegraphics[width=1.7cm,height=3cm]{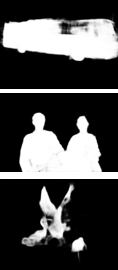}%
\label{fig_csf}}
\hfil
\subfloat[\textbf{DAD \protect\\ ResNet-50}]{
\includegraphics[width=1.7cm,height=3cm]{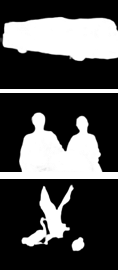}%
\label{fig_dad_r}}
\hfil
\subfloat[\textbf{DAD \protect\\ Res2Net-50}]{
\includegraphics[width=1.7cm,height=3cm]{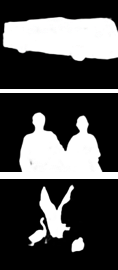}%
\label{fig_dad_r2}}
\hfil
\subfloat[\textbf{DAD \protect\\ PVT-v2-b2}]{
\includegraphics[width=1.7cm,height=3cm]{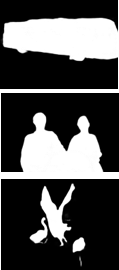}%
\label{fig_dad_p}}
\hfil

\caption{Visual SOD results of different methods.}
\label{fig_sod}
\end{figure*}

\begin{table*}[]
\normalsize
\caption{Performance comparison with baselines on six SOD datasets. The best score of each backbone is highlighted in bold, and the best score for each metric is marked in red. }
\vspace{-0.5em}
\label{tab:table1}
\center
\begin{tabular}{c|c|cc|cc|cc|cc|cc|cc}
\hline
\multirow{2}{*}{Baseline} & \multirow{2}{*}{Backbone}&
  \multicolumn{2}{c|}{DUT-OMRON} &
  \multicolumn{2}{c|}{SOD} &
  \multicolumn{2}{c|}{ECSSD} &
  \multicolumn{2}{c|}{DUTS-TE} &
  \multicolumn{2}{c|}{HKU-IS} &
  \multicolumn{2}{c}{PASCAL-S} \\ \cline{3-14}
       & & $E_{\phi}\uparrow $ &$M\downarrow$ &   $E_{\phi}\uparrow $& $M\downarrow$ &
        $E_{\phi}\uparrow$& $M\downarrow$ &
         $E_{\phi}\uparrow $& $M\downarrow$ &
          $E_{\phi}\uparrow $& $M\downarrow$ &
           $E_{\phi}\uparrow $& $M\downarrow$ \\ \hline
CPD & VGG-16
&0.845  &0.057
&0.787  &0.113
&0.938  &0.040
&0.902  &0.043
&- &-
&0.883  &0.072 \\
EGNet & VGG-16
&0.848  &0.056
&0.802  &0.110
&0.936  &0.041
&0.898  &0.044
&- &-
&0.878  &0.077\\
MINet & VGG-16
&0.841 &0.057
&- &-
&0.941  &0.037
&0.907  &{\textbf{0.039}}
&- &-
&0.891  &0.065\\
GateNet & VGG-16
&0.836  &0.061
&- &-
&0.931  &0.042
&0.893  &0.045
&- &-
&0.891  &0.065\\
PoolNet+ & VGG-16
&0.851  &0.056
&0.815  &0.105
&0.936  &0.042
&0.904  &0.040
&0.940  &0.033
&0.876  &0.075 \\
$\textbf{DAD}$ & VGG-16
&{\textbf{0.869}}  &{\textbf{0.055}}
&{\textbf{0.829}}  &{\textbf{0.098}}
&{\textbf{0.951}}  &{\textbf{0.033}}
&{\textbf{0.919}}  &{\textbf{0.039}}
&{\textbf{0.951}}  &{\textbf{0.030}}
&{\textbf{0.897}}  &{\textbf{0.064}}\\\hline
CPD & ResNet-50
&0.845 &0.057
&0.787 &0.113
&0.938 &0.040
&0.902 &0.043
&- &-
&0.883 &0.072\\
PoolNet & ResNet-50
&0.854 &0.056
&0.818 &0.102
&0.940 &0.039
&0.904 &0.040
&0.940 &0.033
&0.876 &0.075 \\
BANet & ResNet-50
&0.861 &0.061
&0.813 &0.109
&0.940 &0.041
&0.897 &0.046
&0.938 &0.037
&0.875 &0.078\\
EGNet & ResNet-50
&0.848 &0.053
&0.820 &0.097
&0.943 &0.037
&0.907 &0.039
&- &-
&0.881 &0.074\\
MINet & ResNet-50
&0.855 &0.056
&- &-
&0.948 &0.034
&0.917 &0.037
&- &-
&0.893 &0.064\\
GateNet & ResNet-50
&0.851 &0.055
&- &-
&0.934 &0.041
&0.906 &0.040
&- &-
&0.884  &0.068\\
PoolNet+ & ResNet-50
&0.841 &0.054
&0.805 &0.104
&0.945 &0.035
&0.910 &0.037
&- &-
&0.897 &0.065\\
DFI & ResNet-50
&0.864 &0.055
&0.812 &0.102
&0.924 &0.035
&0.892 &0.039
&0.951 &0.031
&0.863 &0.066\\
$\textbf{DAD}$& ResNet-50
&\textbf{0.867} &{\textbf{0.052}}
&{\textbf{0.825}} &{\textbf{0.095}}
&{\textbf{0.953}} &{\textbf{0.032}}
&{\textbf{0.925}} &{\textbf{0.035}}
&{\textbf{0.953}} &{\textbf{0.028}}
&{\textbf{0.901}} &{\textbf{0.060}}\\\hline
CSF  & Res2Net-50
&0.868 &0.057
&0.822 &0.102
&0.947 &0.034
&0.907 &0.042
&0.948 &0.030
&0.878 &0.072\\
PoolNet \# & Res2Net-50
&0.857 &0.053
&0.815 &0.100
&0.946 &0.036
&0.902 &0.041
&0.942 &0.032
&0.884 &0.069\\
$\textbf{DAD}$& Res2Net-50
&{\textbf{0.882}} &{\textbf{0.049}}
&{\textbf{0.834}} &{\textbf{0.091}}
&{\textbf{0.959}} &{\textbf{0.028}}
&{\textbf{0.934}} &{\textbf{0.032}}
&{\textbf{0.958}} &{\textbf{0.026}}
&{\textbf{0.909}} &{\textbf{0.058}}
  \\ \hline
PVT-SOD & PVT-v2-b2
&0.883 &0.044
&- &-
&0.958 &0.028
&0.933 &0.030
&0.960 &0.026
&0.906 &0.057 \\
\textbf{DAD} & PVT-v2-b2
&{\color{red}\textbf{0.903}} &{\color{red}\textbf{0.045}}
&{\color{red}\textbf{0.859}} &{\color{red}\textbf{0.082}}
&{\color{red}\textbf{0.965}} &{\color{red}\textbf{0.024}}
&{\color{red}\textbf{0.949}} &{\color{red}\textbf{0.026}}
&{\color{red}\textbf{0.967}} &{\color{red}\textbf{0.022}}
&{\color{red}\textbf{0.918}} &{\color{red}\textbf{0.052}}
  \\ \hline
\end{tabular}
{\textbf{\#} means that the results were obtained by ourselves.}
\vspace{-1.0em}
\end{table*}

\subsection{Experimental Setting}
In this study, we aimed to design a unified decoder network to capture and enlarge the difference between the foreground and the background, for binary segmentation. To evaluate the proposed model, experiments were conducted in SOD, COD, polyp segmentation, and mirror detection. All the models were implemented in PyTorch 1.7.0, and the training and testing were achieved using an NVIDIA GeForce RTX 3090 GPU with 24 GB memory. The spatial size of the input images was resized to $416\times416$. The Adam optimizer\cite{kingma2014adam} was used in the training process. The initial learning rate was $10^{-4}$, which was attenuated by 10 times every 50 epochs. It is also worth mentioning that the difference-aware decoder was trained for 200 epochs and the batch size was 36.

\subsection{Evaluation Metrics}
Following the previous works, we utilize the  structure-measure ($S_\alpha$) \cite{fan2017structure}, E-measure ($E_{\phi}$) \cite{fan2021cognitive}, weighted F-measure ($F^w_{\beta}$) \cite{margolin2014evaluate} and mean absolute error ($M$) as the evaluation metrics. Structure-measure ($S_\alpha$) \cite{fan2017structure} can evaluate region-aware and object-aware structural similarity between predict map and ground truth, which can be calculated as:
\begin{equation}
	\label{structure-measure}
	\begin{split}
		S_\alpha = \alpha \times S_0 +(1-\alpha) \times S_r,
	\end{split}
\end{equation}
where $\alpha \in[0,1] $, which was set to $0.5$. $S_0$ and $S_r$ represent the region-aware and object-aware structural similarity, respectively. The E-measure ($E_{\phi}$) \cite{fan2021cognitive} can capture the image-level statistics and pixel-level matching information at the same time. It can be calculated as:
\begin{equation}
	\label{E-measure}
	\begin{split}
E_{\phi} =\frac{1}{w \times h} \sum_{x=1}^{w} \sum_{y=1}^{h} \phi_{\mathrm{FM}}(x, y),
	\end{split}
\end{equation}
where $\phi_{\mathrm{FM}}$ is the enhanced alignment matrix, and $w$ and $h$ respectively represent the width and height of the image. Weighted F-measure ($F^w_{\beta}$) \cite{margolin2014evaluate} evaluates the segmentation result by considering each pixel independently. It can be calculated as:
\begin{equation}
	\label{Weighted F-measure}
	\begin{split}
		F^w_{\beta} =\frac{\left(1+\beta^{2}\right) \text { precision }^{\omega} \cdot \text { recall }^{\omega}}{\beta^{2} \cdot \text { precision }^{\omega}+\text { recall }^{\omega}}.
	\end{split}
\end{equation}
Mean absolute error ($M$) calculates the error between the predicted map $S$ and the ground truth $G$. It is formulated as:
\begin{equation}
	\label{Mean absolute error}
	\begin{split}
	M =\frac{1}{W \times H} \sum_{x=1}^{W} \sum_{y=1}^{H}\|S(x, y)-G(x, y)\|
	\end{split},
\end{equation}
where $W$ and $H$ respectively represent the width and height of the image.
\par
For both the SOD and COD tasks, we selected the structure-measure ($S_\alpha$) \cite{fan2017structure}, E-measure ($E_{\phi}$) \cite{fan2021cognitive}, weighted F-measure ($F^w_{\beta}$) \cite{margolin2014evaluate} and mean absolute error ($M$). For the polyp segmentation, we not only selected the structure-measure ($S_\alpha$) \cite{fan2017structure}, weighted F-measure ($F^w_{\beta}$) \cite{margolin2014evaluate} and mean absolute error ($M$), but we also selected the mean value of the dice coefficients\cite{milletari2016v} and $IoU$, which mainly focus on the internal consistency of the segmentation results.

\begin{figure*}[t]
\centering
\subfloat[Image \protect\\ Backbone ]{
\includegraphics[width=1.7cm,,height=5cm]{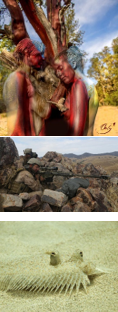}%
\label{fig_codimage}}
\hfil
\subfloat[GT]{
\includegraphics[width=1.7cm,,height=5cm]{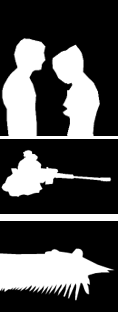}%
\label{fig_codgt}}
\hfil
\subfloat[LSR \protect\\  ResNet-50]{
\includegraphics[width=1.7cm,,height=5cm]{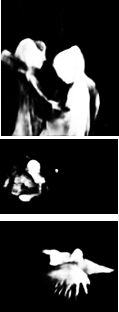}%
\label{fig_lsr}}
\hfil
\subfloat[MGL \protect\\  ResNet-50]{
\includegraphics[width=1.7cm,,height=5cm]{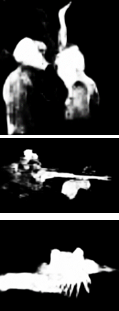}%
\label{fig_MGL-R}}
\hfil
\subfloat[PFNet \protect\\ ResNet-50]{
\includegraphics[width=1.7cm,,height=5cm]{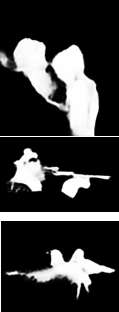}%
\label{fig_pfnet}}
\hfil
\subfloat[SINetV2 \protect\\ Res2Net-50]{
\includegraphics[width=1.7cm,,height=5cm]{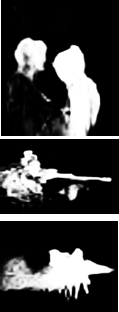}%
\label{fig_sinetv2}}
\hfil
\subfloat[SINetV2 \protect\\ PVT-b2-v2]{
\includegraphics[width=1.7cm,,height=5cm]{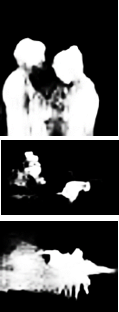}%
\label{fig_sinetv2_P}}
\hfil
\subfloat[\textbf{DAD \protect\\ ResNet-50}]{
\includegraphics[width=1.7cm,,height=5cm]{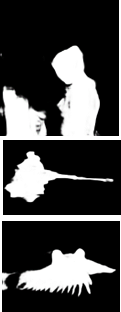}%
\label{fig_dad_r}}
\hfil
\subfloat[\textbf{DAD \protect\\ Res2Net-50}]{
\includegraphics[width=1.7cm,,height=5cm]{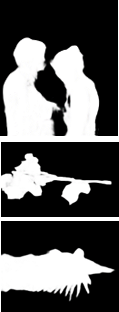}%
\label{fig_DAd_r2}}
\hfil
\subfloat[\textbf{DAD \protect\\ PVT-b2-v2}]{
\includegraphics[width=1.7cm,,height=5cm]{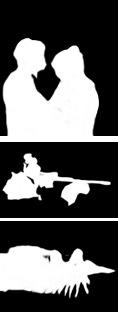}%
\label{fig_dad_p}}
\hfil
\caption{Visual COD results of different methods.}
\label{fig_cod}
\end{figure*}

\begin{table*}[]
\setlength\tabcolsep{3pt}
\caption{Performance comparison with baseline models on COD datasets. $\uparrow$ indicates the higher the score the better and vice versa. The best score of each backbone is highlighted in bold, and the best score for each metric is marked in red.}
\center
\label{tab:1}
\normalsize
\begin{tabular}{c|c|cccc|cccc|cccc|cccc}
\hline
\multirow{2}{*}{Baseline}& \multirow{2}{*}{Backbone} & \multicolumn{4}{c|}{CAMO}&\multicolumn{4}{c|}{CHAMELLON} &\multicolumn{4}{c|}{COD10K} &\multicolumn{4}{c}{NC4K}                                                 \\ \cline{3-18}
 & & $S_{\alpha}\uparrow $ & $E_{\phi}\uparrow$ & $F^w_{\beta}\uparrow$ & $M\downarrow$ & $S_{\alpha}\uparrow $ & $E_{\phi}\uparrow$ & $F^w_{\beta}\uparrow$ & $M\downarrow$& $S_{\alpha}\uparrow $ & $E_{\phi}\uparrow$ & $F^w_{\beta}\uparrow$ & $M\downarrow$ & $S_{\alpha}\uparrow $ & $E_{\phi}\uparrow$ & $F^w_{\beta}\uparrow$ & $M\downarrow$ \\ \hline
SINet & ResNet-50
& 0.751 & 0.771 & 0.606 & 0.100
& 0.869 & 0.891 & 0.740 & 0.044
& 0.771 & 0.806 & 0.551 & 0.051
& 0.808 & 0.871 & 0.738 & 0.058\\
LSR & ResNet-50
& 0.787 & 0.838 & 0.696 & 0.080
& 0.890 & 0.935 & {\textbf{0.822}} & 0.030
& 0.804 & 0.877 & 0.660 & 0.040
& {\textbf{0.839}} & 0.883 & {\textbf{0.777}} & 0.053\\
MGL-R & ResNet-50
& 0.775 & 0.812 & 0.673 & 0.088
& {\textbf{0.893}} & 0.917 & 0.808 & 0.031
& 0.814 & 0.851 & {\textbf{0.666}} & {\textbf{0.035}}
& 0.833 & 0.866 & 0.754 & 0.053\\
MGL-S & ResNet-50
& 0.772 & 0.806 & 0.664 & 0.089
& 0.892 & 0.911 & 0.799 & 0.032
& 0.811 & 0.844 & 0.654 & 0.037
& 0.827 & 0.860 & 0.747 & 0.055\\
UGTR & ResNet-50
& 0.784 & 0.821 & 0.683 & 0.086
& 0.888 & 0.910 & 0.787 & 0.031
& {\textbf{0.817}} & 0.852 & 0.665 & 0.036
& {\textbf{0.839}} & 0.874 & 0.755 & 0.052\\
PFNet & ResNet-50
& 0.782 & 0.842 & 0.695 & 0.085
& 0.882 & 0.931 & 0.808 & 0.032
& 0.800 & 0.877 & 0.660 & 0.040
& 0.829 & 0.886 & 0.747 & 0.053\\
EINet & ResNet-50
&0.626 &0.675 &0.424 &0.143
&0.793 &0.850 &0.631 &0.069
&0.636 &0.708 &0.363 &0.090
&0.676 &0.735 &0.483 &0.112 \\
$\textbf{DAD}$& ResNet-50
& {\textbf{0.795}} & {\textbf{0.863}} & {\textbf{0.713}} & {\textbf{0.076}}
& 0.885 & {\textbf{0.941}} & 0.801 & {\textbf{0.028}}
& 0.803 & {\textbf{0.886}} & 0.659 & 0.038
& 0.834 & {\textbf{0.900}} & 0.758 & {\textbf{0.049}} \\ \hline
SINet V2 & Res2Net-50
& 0.820 & 0.882 & 0.743 & 0.070
& 0.888 & 0.942 & 0.816 & 0.030
& 0.815 & 0.887 & 0.680 & 0.037
& 0.847 & 0.903 & 0.767 & 0.048\\
EINet & Res2Net-50
&0.817 &0.872 &0.740 &0.070
&0.891 &0.939 &0.819 &0.030
&0.815 &0.887 &0.682 &0.036
&0.845 &0.900 &0.768 &0.047 \\
$\textbf{DAD}$  & Res2Net-50
& {\textbf{0.830}} & {\textbf{0.895}} & {\textbf{0.774}} & {\textbf{0.063}}
& {\textbf{0.899}} & {\textbf{0.947}} & {\textbf{0.842}} & {\textbf{0.027}}
& {\textbf{0.827}} & {\textbf{0.905}} & {\textbf{0.720}} & {\textbf{0.032}}
& {\textbf{0.851}} & {\textbf{0.911}} & {\textbf{0.792}} & {\textbf{0.044}}\\ \hline
SINet V2 \#& PVT-v2-b2
& 0.863 & 0.918 & 0.791 & 0.051
& 0.892 & 0.946 & 0.810 & 0.028
& 0.847 & 0.915 & 0.719 & 0.028
& 0.878 & 0.928 & 0.811 & 0.036\\
DIIT  & PVT-v2-b2
&0.857 &0.916  &0.796  &0.050
&- &-  &-  &-
&0.824 &0.896  &0.695  &0.034
&0.863 &0.917  &0.792  &0.041\\
EINet & PVT-v2-b2
&0.856 &0.910 &0.801 &0.054
&0.895 &0.944 &0.837 &0.027
&0.847 &0.915 &0.742 &0.028
&0.875 &0.926 &0.817 &0.037 \\
$\textbf{DAD}$  & PVT-v2-b2
&{\color{red}\textbf{0.867}} &{\color{red}\textbf{0.929}}  &{\color{red}\textbf{0.821}}  &{\color{red}\textbf{0.047}}
&{\color{red}\textbf{0.905}} &{\color{red}\textbf{0.963}}  &{\color{red}\textbf{0.855}}  &{\color{red}\textbf{0.022}}
&{\color{red}\textbf{0.864}} &{\color{red}\textbf{0.935}}  &{\color{red}\textbf{0.776}}  &{\color{red}\textbf{0.023}}
&{\color{red}\textbf{0.882}} &{\color{red}\textbf{0.935}}  &{\color{red}\textbf{0.839}}  &{\color{red}\textbf{0.033}}\\ \hline
\end{tabular}
{\textbf{\#} means that the results were obtained by ourselves.}
\vspace{-1.0em}
\end{table*}

\subsection{Experiments in SOD}
\subsubsection{Datasets}
We kept the same experimental setup as used for other SOD methods\cite{DBLP:journals/corr/abs-2101-04704}. The training dataset had 10,553 images, and we conducted experiments on six datasets. The DUT-OMRON dataset\cite{yang2013saliency} has 5,168 images with complex objects. The PASCAL-S dataset\cite{li2014secrets} consists of 850 challenging images. The HKU-IS dataset \cite{li2015visual} contains 4,447 images with multiple foreground objects. The Extended Complex Scene Saliency Dataset (ECSSD)\cite{yan2013hierarchical} and Salient Objects Dataset (SOD)\cite{movahedi2010design} contain 1,000 images and 300 images, respectively. The DUTS dataset\cite{wang2017learning} is made up of the DUTS-TR (10,553 images for training) and DUTS-TE (5,019 images for testing) sets. In the experiments, the DUTS-TR set was used for the training, and the DUTS-TE set was used for the testing.

\subsubsection{Compared Methods}
For the SOD task, we selected 16 methods to evaluate the proposed difference-aware decoder: methods based on the VGG-16 backbone, $i.e.$,
CPD (VGG-16)\cite{wu2019cascaded},
EGNet (VGG-16)\cite{zhao2019egnet},
MINet (VGG-16)\cite{pang2020multi},
GateNet (VGG-16)\cite{zhao2020suppress},
PoolNet + (VGG-16)\cite{Liu21PamiPoolNet};
methods based on the ResNet-50 backbone, $i.e.$,
CPD (ResNet-50)\cite{wu2019cascaded},
PoolNet (ResNet-50)\cite{liu2019simple},
BANet (ResNet-50)\cite{su2019selectivity},
EGNet (ResNet-50)\cite{zhao2019egnet},
MINet (ResNet-50)\cite{pang2020multi},
GateNet (ResNet-50)\cite{zhao2020suppress},
PoolNet + (ResNet-50)\cite{Liu21PamiPoolNet},
DFI (ResNet-50) \cite{liu2020dynamic};
methods based on the Res2Net-50 backbone, $i.e.$,
CSF (Res2Net-50) \cite{gao2020highly},
PoolNet (Res2Net-50) \cite{Liu21PamiPoolNet};
and a method based on the PVT-v2-b2 backbone, $i.e.$, PVT-SOD (PVT-v2-b2)\cite{xu2022semantic}. We directly used the results or the code provided by the related authors for the comparison. The proposed difference-aware decoder was tested on the VGG-16, ResNet-50, Res2Net-50, and PVT-v2-b2 backbones, for a fair comparison. It is worth noting that PVT-v2-b2 only has a four-layer encoder structure. Therein, for the proposed DAD, we select the first and fourth layer features for Stage A; whereas the first, second and third layers for Stage B.

\subsubsection{Visual Results of the SOD}
In order to compare the visual effects of the different SOD methods, several salient images are selected for visualization in Fig. \ref{fig_sod}. Compared with the other state-of-the-art SOD methods, the difference-aware decoder achieves the best segmentation effects. For the first image, the biggest difficulty is the segmentation of the bus roof from the sky, which are highly mixed. The compared methods of GateNet, MINet, and CSF fail to successfully segment the bus roof from the sky. The PoolNet and PoolNet+ methods can somehow deal with this problem, but there are still some flaws. The difference-aware decoder, with ResNet-50, Res2Net-50, and PVT-v2-b2 as the backbone, can perfectly distinguish the clear outline of the bus from the background. For the second and the third images, it is apparent that there are ambiguous regions in the segmentation results of the other methods, but the proposed method with different backbones can achieve maps that are much more similar to the ground truth.
\subsubsection{Quantitative Evaluation of the SOD}
Table \ref{tab:table1} lists the overall quantitative evaluation results of the different methods obtained on the six datasets. The best score for each backbone is highlighted in bold, and the best score for each metric is marked in red. It is clear that the proposed difference-aware decoder with PVT-v2-b2 backbone achieves the best results, for all the datasets and evaluation metrics. What is more, with the different backbones of VGG-16, ResNet-50, and Res2Net-50, the difference-aware decoder always achieves the best performance, indicating the transferability of the proposed decoder architecture for different backbones.
\par
Compared to the other methods, the difference-aware decoder improves the evaluation metrics by a large margin. The main reason for this advantage is that the proposed difference-aware decoder can more effectively enlarge the difference between the foreground and background, by fusing the guide map and the background-aware features.

\begin{figure*}[t]
\centering
\subfloat[Image]{
\includegraphics[width=2.5cm,,height=4cm]{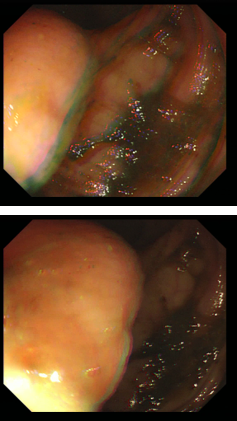}%
\label{fig_polypimage}}
\hfil
\subfloat[GT]{
\includegraphics[width=2.5cm,height=4cm]{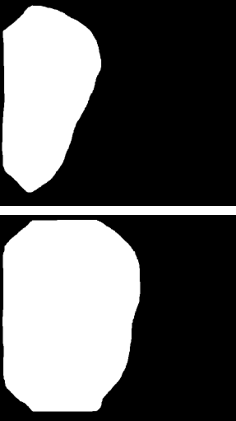}%
\label{fig_polypgt}}
\hfil
\subfloat[PraNet]{
\includegraphics[width=2.5cm,height=4cm]{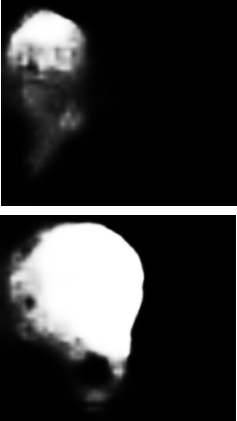}%
\label{fig_pranet}}
\hfil
\subfloat[EUNet]{
\includegraphics[width=2.5cm,height=4cm]{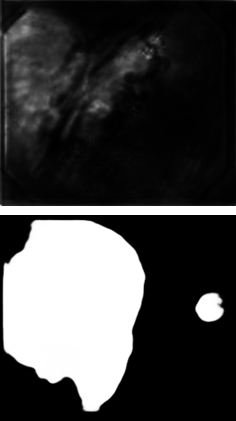}%
\label{fig_eunet}}
\hfil
\subfloat[SANet]{
\includegraphics[width=2.5cm,height=4cm]{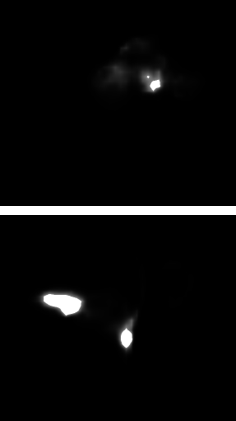}%
\label{fig_sanet}}
\hfil
\subfloat[Polyp-PVT]{
\includegraphics[width=2.5cm,height=4cm]{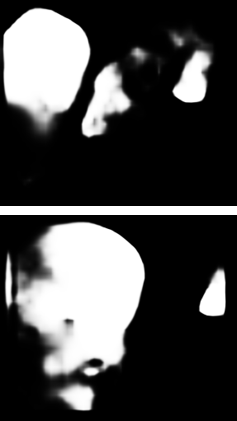}%
\label{fig_UGTR}}
\hfil
\subfloat[\textbf{DAD(ours)}]{
\includegraphics[width=2.5cm,height=4cm]{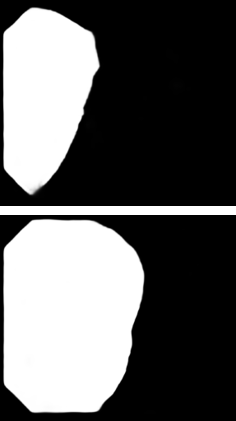}%
\label{fig_DAD_POLYP}}
\hfil
\caption{Visual Polyp Segmentation results of different methods.}
\label{fig_polyp}
\end{figure*}

\begin{table*}[]
\setlength\tabcolsep{3pt}
\caption{Performance comparison with baseline models on polyp segmentation datasets. $\uparrow$ indicates the higher the score the better and vice versa. The best score for each metric is marked in red.}
\center
\label{tab:polyp}
\normalsize
\setlength\tabcolsep{2pt}
\begin{tabular}{c|c|ccccc|ccccc|ccccc}
\hline
\multirow{2}{*}{Baseline} & \multirow{2}{*}{Backbone} & \multicolumn{5}{c|}{ClinicDB}&\multicolumn{5}{c|}{ColonDB} &\multicolumn{5}{c}{ETIS}                                              \\ \cline{3-17}
&  & $Dice\uparrow$ & $IoU\uparrow$ & $F^w_{\beta}\uparrow$ & $S_{\alpha}\uparrow$ & $M\downarrow$
 & $Dice\uparrow$ & $IoU\uparrow$ & $F^w_{\beta}\uparrow$ & $S_{\alpha}\uparrow$ & $M\downarrow$
 & $Dice\uparrow$ & $IoU\uparrow$ & $F^w_{\beta}\uparrow$ & $S_{\alpha}\uparrow$ & $M\downarrow$ \\ \hline
UNet & ResNet-50
&0.823 &0.755 &0.811 &0.889 &0.019
&0.512 &0.444 &0.498 &0.712 &0.061
&0.398 &0.335 &0.366 &0.684 &0.036\\
UNet++ & ResNet-50
&0.794 &0.729 &0.785 &0.873 &0.022
&0.483 &0.410 &0.467 &0.691 &0.064
&0.401 &0.344 &0.390 &0.683 &0.035\\
MSEG & ResNet-50
&0.909 &0.864 &0.907 &0.938 &0.007
&0.735 &0.666 &0.724 &0.834 &0.038
&0.700 &0.630 &0.671 &0.828 &0.015\\
ACSNet &ResNet-50
&0.882 &0.826 &0.873 &0.927 &0.011
&0.716 &0.649 &0.697 &0.829 &0.039
&0.578 &0.509 &0.530 &0.754 &0.059\\
PraNet &Res2Net-50
&0.899 &0.849 &0.896 &0.936 &0.009
&0.712 &0.640 &0.699 &0.820 &0.043
&0.628 &0.567 &0.600 &0.794 &0.031\\
EU-Net& ResNet-34
&0.902 &0.846 &0.891 &0.936 &0.011
&0.756 &0.681 &0.730 &0.831 &0.045
&0.687 &0.609 &0.636 &0.793 &0.067\\
SANet & Res2Net-50
&0.916 &0.859 &0.909 &0.939 &0.012
&0.753 &0.670 &0.726 &0.837 &0.043
&0.750 &0.654 &0.685 &0.849 &0.015\\
Polyp-PVT & PVT-v2-b2
&0.937 &0.889 &0.936 &0.949 &{\color{red}\textbf{0.006}}
&0.808 &0.727 &0.795 &0.865 &0.031
&0.787 &0.706 &0.750 &0.871 &{\color{red}\textbf{0.013}} \\
\textbf{DAD} & PVT-v2-b2
&{\color{red}\textbf{0.940}} &{\color{red}\textbf{0.893}} &{\color{red}\textbf{0.937}} &{\color{red}\textbf{0.954}} &{\color{red}\textbf{0.006}}
&{\color{red}\textbf{0.826}} &{\color{red}\textbf{0.751}} &{\color{red}\textbf{0.809}} &{\color{red}\textbf{0.880}} &{\color{red}\textbf{0.027}}
&{\color{red}\textbf{0.801}} &{\color{red}\textbf{0.726}} &{\color{red}\textbf{0.763}} &{\color{red}\textbf{0.880}} &0.017
\\ \hline
\end{tabular}
\vspace{-0.1em}
\end{table*}

\subsection{Experiments in COD}
\subsubsection{Datasets}
Four datasets were used to evaluate the proposed model: CHAMELEMON \cite{skurowski2018animal}, CAMO \cite{le2019anabranch}, COD10K \cite{fan2020camouflaged}, NC4K\cite{lv2021simultaneously}. These four datasets differ significantly. NC4K is the largest dataset, which contains 4,121 images downloaded from the Internet. COD10K includes 78 camouflaged categories, with 3,040 training images and 2,026 test images in total. CAMO provides 1,250 images in total, including 1,000 training images and 250 test images of eight categories. The smallest dataset is the CHAMELEON dataset, which does not provide a training set, and contains only 76 images for testing. According to \cite{9444794,mei2021camouflaged,fan2020camouflaged}, we used the training images from the CAMO and COD10K datasets as the training set (4,040 images) and the test images from the CHAMELEMON \cite{skurowski2018animal}, CAMO \cite{le2019anabranch}, COD10K \cite{fan2020camouflaged}, and NC4K\cite{lv2021simultaneously} as the test set.

\subsubsection{Compared Methods}
In order to evaluate the proposed difference-aware decoder, we selected 12 state-of-the-art methods for comparison: methods based on the ResNet-50 backbone, $i.e.$, SINet (ResNet-50) \cite{fan2020camouflaged},
LSR (ResNet-50) \cite{lv2021simultaneously},
MGL-R (ResNet-50)\cite{zhai2021mutual},
MGL-S (ResNet-50)\cite{zhai2021mutual},
UGTR (ResNet-50)\cite{yang2021uncertainty},
PFNet (ResNet-50) \cite{Mei_2021_CVPR},
EINet (ResNet-50)\cite{li2022einet}; methods based on the Res2Net-50 backbone, $i.e.$, SINet V2 (Res2Net-50)\cite{9444794} and EINet (Res2Net-50)\cite{li2022einet}; and methods based on the PVT-v2-b2 backbone, $i.e.$, SINet V2 (PVT-v2-b2)\cite{9444794}, EINet (PVT-v2-b2)\cite{li2022einet}, and DTIT (PVT-v2-b2) \cite{liu2022boosting}. The proposed difference-aware decoder was tested on the ResNet-50, Res2Net-50, and PVT-v2-b2 backbones, for a fair comparison.

\subsubsection{Visual Results of the COD}
In order to compare the visual effects of the different methods, we selected several challenging camouflaged images for the visualization, which are shown in Fig. \ref{fig_cod}. In the first image, the left person is perfectly hidden in the trees. The LSR, PFNet, and SINet V2 methods and the proposed difference-aware decoder with ResNet-50 backbone fail to segment the body of the left person. However, the difference-aware decoder with Res2Net-50 and PVT-b2-v2 backbones can clearly detect both persons. For the second image, the two soldiers are hidden in the rocks, and the other compared methods cannot segment them accurately, but the difference-aware decoder with different backbones can find them successfully. For the third image, the creatures are mixed with their surroundings and are difficult to distinguish. The segmentation results achieved by the compared methods are incomplete, referring to the ground truth, whereas the difference-aware decoder with Res2Net-50 and PVT-b2-v2 backbones can achieve much better segmentation maps. From our understanding, the backbone is important for the COD task. However, the proposed decoder paradigm can make the best use of different backbones to achieve a superior accuracy.

\subsubsection{Quantitative Evaluation of the COD}
Table \ref{tab:1} lists the overall quantitative evaluation results of the different methods obtained on the four datasets. On the ResNet-50 backbone, the proposed difference-aware decoder achieves the best score in the $E_{\phi}$ metric for all four datasets. With regard to the $S_{\alpha}$ and $F^w_{\beta}$ metrics, the difference-aware decoder achieves the best scores on the CAMO dataset and the second-best scores on the NC4K dataset. With regard to the $M$ metric, the difference-aware decoder achieves the best scores on the CAMO, CHAMELEON, and NC4K datasets. For the COD task, Res2Net-50 and PVT-v2-b2 are the better backbones\cite{fan2020camouflaged,liu2022boosting}. On these two backbones, the proposed difference-aware decoder outperforms SINet V2 and DTIT by a large margin, for all the metrics and datasets. The main reason for this advantage is that the proposed difference-aware decoder can more effectively enlarge the difference between the camouflaged objects and background, by fusing the guide map and the background-aware features.

\subsection{Experiments in Polyp Segmentation}
\subsubsection{Datasets}
According to the experimental settings used for PraNet \cite{fan2020pranet}, we tested the trained model on three public datasets: ClinicDB\cite{bernal2015wm}, ColonDB\cite{tajbakhsh2015automated}, and ETIS\cite{silva2014toward}. The recently proposed Polyp-PVT\cite{dong2021polyp} utilizes PVT-v2-b2\cite{wang2021pyramid} as the backbone, and shows a great performance in polyp segmentation.

\subsubsection{Compared Methods}
In order to evaluate the proposed difference-aware decoder, we selected eight state-of-the-art methods for comparison: UNet\cite{ronneberger2015u}, UNet++\cite{zhou2018unet++}, MSEG\cite{huang2021hardnet}, ACSNet\cite{zhang2020adaptive}, PraNet\cite{fan2020pranet}, EU-Net\cite{patel2021enhanced}, SANet\cite{wei2021shallow}, and Polyp-PVT\cite{dong2021polyp}. Following Polyp-PVT, the difference-aware decoder also used PVT-v2-b2 as the backbone for the experiments.

\subsubsection{Visual Results of the Polyp Segmentation}
In Fig. \ref{fig_polyp}, two examples are selected for visualization. From these two examples, it can be seen that the polyps are highly mixed with the background. The compared methods of PraNet, EUNet, and SANet fail to achieve efficient segmentation, and Polyp-PVT can achieve only a rough map. However, the difference-aware decoder can segment the complete outline of the polyps successfully, compared to the ground truth.

\subsubsection{Quantitative Evaluation of Polyp Segmentation}
Table \ref{tab:polyp} lists the overall quantitative evaluation results of the different methods obtained on the three datasets. For the $Dice$, $IoU$, $F^w_{\beta}$, and $S_{\alpha}$ metrics, the difference-aware decoder achieves the best performance on the ClinicDB, ColonDB, and ETIS datasets. For the $M$ metric, the difference-aware decoder obtains the best accuracy on the ClinicDB and ColonDB datasets. These quantitative evaluation results reflect the effectiveness of the proposed decoder architecture in the polyp segmentation task.
\begin{figure}[]
\vspace{-1.0em}
\centering
\subfloat[Image]{
\includegraphics[width=1.6cm,height=6cm]{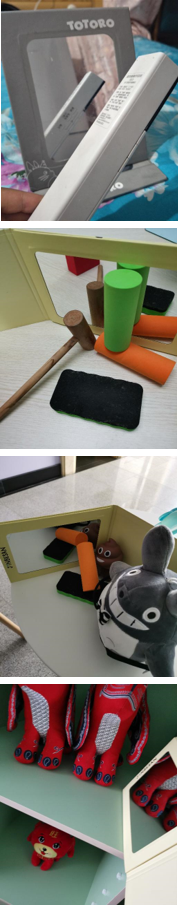}%
\label{fig_msd_image_sup}}
\hfil
\subfloat[GT]{
\includegraphics[width=1.6cm,height=6cm]{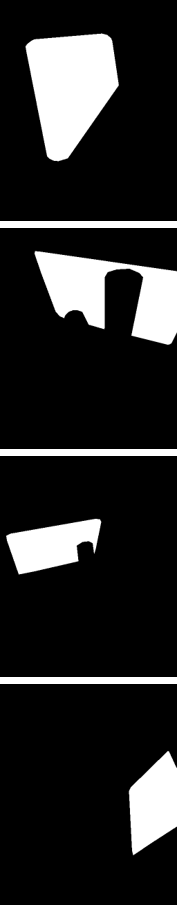}%
\label{fig_msd_gt_sup}}
\hfil
\subfloat[MirrorNet]{
\includegraphics[width=1.6cm,height=6cm]{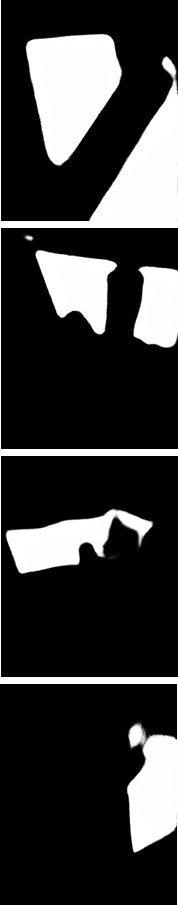}%
\label{fig_mirrornet_sup}}
\hfil
\subfloat[LSA]{
\includegraphics[width=1.6cm,height=6cm]{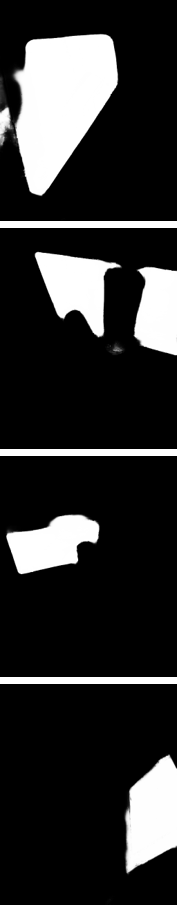}%
\label{fig_lsa_sup}}
\hfil
\subfloat[\textbf{DAD}]{
\includegraphics[width=1.6cm,height=6cm]{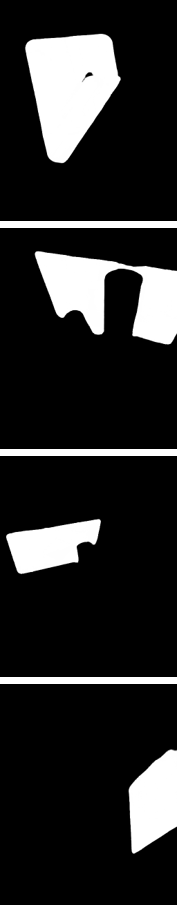}%
\label{fig_dad_p_sup}}
\hfil
\caption{Visual results of the different methods obtained on the MSD dataset.}
\label{fig_mirror_sup}
\vspace{-1.0em}
\end{figure}
\subsection{Experiments in Mirror Detection}
\subsubsection{Datasets}
The MSD\cite{yang2019my} dataset was chosen for the mirror detection experiments. The MSD dataset consists of 4,018 pairs of images, with 3,063 pairs of images for training and 955 pairs of images for testing.
\subsubsection{Compared Methods}
In order to evaluate the proposed difference-aware decoder, we selected two general segmentation methods: PSPNet\cite{zhao2017pyramid}, and DeepLab v3+\cite{chen2018encoder};
and three state-of-the-art methods designed for mirror detection: MirrorNet\cite{yang2019my}, PMD-Net\cite{lin2020progressive}, and LSA\cite{Guan_2022_CVPR}.
For a fair comparison, we used ResNeXt-101 \cite{xie2017aggregated} pre-trained on
ImageNet-1K \cite{deng2009imagenet} as the backbone for all the methods.
Conditional random fields (CRF)\cite{krahenbuhl2011efficient} was also utilized as a common post-processing method for all the methods during the inference.
\subsubsection{Visual Results of the Mirror Detection}
Fig. \ref{fig_mirror_sup} illustrates several examples for the visual comparison between the proposed method and the other state-of-the-art methods. In these examples, it can be found that the backgrounds of the mirrors are complex and it is difficult for the other methods to segment relatively clear bodies for the mirrors. In contrast, the difference-aware decoder can achieve a segmentation map that is closer to the ground truth, showing the superiority of the proposed method.

\subsubsection{Quantitative Evaluation of the Mirror Detection}
Following the evaluation strategy used for other methods\cite{Guan_2022_CVPR}, we used the F1-score ($F1$), Intersection over Union ($IoU$), accuracy ($Acc$), and $M$ metric to measure the performance of the proposed difference-aware decoder and the other compared methods.
From Table \ref{tab:table_mirror_all}, it can be found that the proposed method achieves the best score in all the metrics.

\begin{table}[]
\normalsize
\setlength\tabcolsep{3pt}
\caption{Performance comparison with baseline models on MSD dataset. $\uparrow$ indicates the higher the score the better and vice versa. The best score for each metric is marked in red.}
\label{tab:table_mirror_all}
\center
\begin{tabular}{c|c|cccc}
\hline
\multirow{2}{*}{Baseline} & \multirow{2}{*}{Backbone}&
  \multicolumn{4}{c}{MSD dataset} \\ \cline{3-6}
      & & $F1\uparrow$ & $IoU\uparrow$ & $Acc\uparrow$ & $M\downarrow$ \\ \hline
PSPNet& ResNext-101 &
0.8459 &67.99 &92.19 &0.07875\\
Deeplab v3+& ResNext-101 &
0.8750 &77.48 &94.13 &0.05932\\
MirrorNet& ResNext-101 &
0.8597 &77.41 &92.75 &0.07257 \\
PMD-Net& ResNext-101 &
0.8691 &76.88 &93.94 &0.06130\\
LSA & ResNext-101 &
0.8887 &79.85 &94.63 &0.05421 \\ \hline
\textbf{DAD} & ResNext-101 &
{\color{red}\textbf{0.8910}} &{\color{red}\textbf{82.90}} &{\color{red}\textbf{95.37}} &{\color{red}\textbf{0.04644}} \\ \hline
\end{tabular}
\vspace{-1.0em}
\end{table}

\subsection{Parametric Analysis}
In order to verify the effectiveness of the parameter selection in each module in the experiments, we used Res2Net-50 as the backbone and selected two COD datasets (CAMO \cite{le2019anabranch}, COD10K \cite{fan2020camouflaged}) and two SOD datasets (ECSSD\cite{yan2013hierarchical}, PASCAL-S\cite{li2014secrets}) for the validation experiments.

\subsubsection{Multi-Layer Feature Partition}
\begin{table}[htb]
\centering
\vspace{-1.0em}
\caption{Multi-layer feature partition analysis for the GMG (Stage A) and MFF module (Stage B).}
\label{tab:4}
\small
\setlength\tabcolsep{2pt}
\begin{tabular}{c|cc|cc|cc|cc}
\hline
\multirow{2}{*}{Layers}
& \multicolumn{2}{c|}{CAMO} & \multicolumn{2}{c|}{COD10K} & \multicolumn{2}{c|}{ECSSD} & \multicolumn{2}{c}{PASCAL-S}
\\ \cline{2-9}
& $E_{\phi}\uparrow$ & $M\downarrow$
& $E_{\phi}\uparrow$ & $M\downarrow$
& $E_{\phi}\uparrow$ & $M\downarrow$
& $E_{\phi}\uparrow$ & $M\downarrow$  \\ \hline
2+5
& 0.890 & 0.069
& 0.904 & 0.032
& 0.951 & 0.033
& 0.903 & 0.061\\
3+5
& 0.894 & 0.065
& 0.901 & 0.033
& 0.947 & 0.031
& 0.899 & 0.062\\
4+5
& 0.876 & 0.072
& 0.882 & 0.038
& 0.957 & 0.029
& 0.902 & 0.060 \\
5
& 0.865 & 0.079
& 0.875 & 0.042
& 0.927 & 0.037
& 0.888 & 0.069\\
1+2+5
& 0.880 & 0.071
&\textbf{0.909} & 0.032
& 0.953 & 0.030
& 0.900 & 0.059\\
1+3+5
& 0.887 & 0.065
& 0.905 & \textbf{0.031}
& 0.955 & 0.031
& 0.902 & 0.060\\
1+4+5
& 0.883 & 0.069
& 0.904 & 0.032
& 0.953 & 0.029
& 0.897 & 0.061\\
1+2+4+5
& 0.885 & 0.071
& 0.895 & 0.035
& 0.954 & 0.029
& 0.895 & 0.060\\
1+2+3+5
& 0.888 & 0.068
& 0.899 & 0.033
& 0.952 & 0.030
& 0.902 & 0.060\\
1+3+4+5
& 0.880 & 0.072
& 0.904 & 0.032
& 0.954 & 0.029
& 0.902 & 0.059
\\ \hline
1+5(proposed) & \textbf{0.895} & \textbf{0.063}
& 0.905 & 0.032
& \textbf{0.963} & \textbf{0.028}
& \textbf{0.909} & \textbf{0.058} \\ \hline
\end{tabular}
\end{table}
In Stage A, we chose layers 1 and 5 from the backbone for the guide map generation; and in Stage B, we fused layers 2~4 for the background-aware features extraction. In the following, we attempt to explain why we divided the multi-layer features in this way. From experience \cite{chen2017deeplab,fan2020camouflaged}, the highest-level features are always used for the coarse map generation. For a complete analysis, we combined layer 5 with different layer(s) to finish Stage A, and the other layer(s) for Stage B. Table \ref{tab:4} presents the results of the different combinations of multi-layer features. It can be clearly observed that the strategy of combining layers 1 and 5 for Stage A can achieve the best accuracy on the CAMO, ECSSD, and PASCAL-S datasets, while almost achieving the best results on the COD10K dataset. Therefore, in the experiments, we fixed layers 1 and 5 for Stage A. The explanation for this is that the high resolution of layer 1 can make up more spatial details for layer 5, which is an extremely effective way to obtain a fine guide map. It needs to be clarified that the multi-layer features were divided without overlapping. This is mainly done to enhance the differences between the guide map from Stage A and the background-aware features from Stage B.

\subsubsection{Spatial Size in Middle Feature Fusion (MFF)}
\begin{table}[htb]
\centering
\caption{Spatial size analysis for the MFF module. B-U is the abbreviation for the Bottom-Up, and T-D is the abbreviation for the Top-Down. }
\label{tab:5}
\small
\setlength\tabcolsep{2pt}
\begin{tabular}{c|cc|cc|cc|cc}
\hline
\multirow{2}{*}{Fusion}
& \multicolumn{2}{c|}{CAMO} & \multicolumn{2}{c|}{COD10K} & \multicolumn{2}{c|}{ECSSD} & \multicolumn{2}{c}{PASCAL-S}
\\ \cline{2-9}
& $E_{\phi}\uparrow$ & $M\downarrow$
& $E_{\phi}\uparrow$ & $M\downarrow$
& $E_{\phi}\uparrow$ & $M\downarrow$
& $E_{\phi}\uparrow$ & $M\downarrow$  \\ \hline
B-U
& 0.885 & 0.071
& 0.903 & 0.034
& 0.957 & 0.029
& 0.901 & 0.059\\
T-D
& 0.886 & 0.068
& 0.901 & 0.033
& 0.952 & 0.031
& 0.905 & 0.060\\
JPU\cite{wu2019fastfcn}
& 0.884 & 0.071
& 0.900 & 0.034
& 0.953 & 0.029
& 0.904 & 0.060
\\ \hline
M(proposed)
& \textbf{0.895} & \textbf{0.063}
& \textbf{0.905} & \textbf{0.032}
& \textbf{0.963} & \textbf{0.028}
& \textbf{0.909} & \textbf{0.058} \\ \hline
\end{tabular}
\end{table}
In Stage B, the proposed MFF module was used to integrate the features at different levels. In fact, the upsample/downsample function is utilized to resize layers 2 and 4 to the size of layer 3. In Table 5, Top-Down means that layer 2 and layer 3 were downsampled to the same size as layer 4, while Bottom-Up means that layer 3 and layer 4 are upsampled to the size of layer 4. Table \ref{tab:5} indicates that the proposed strategy can achieve the best detection accuracy. This is mainly because, if the resolution is low, the proportion of the features that relates to the difference is also reduced. In addition, if resolution is blindly pursued, it would result in the loss of semantic information and an increase in the computational complexity. The joint pyramid upsampling (JPU\cite{wu2019fastfcn}) method also takes the three feature maps as the inputs and utilizes a context module to generate a high-resolution feature map, which is similar to the MFF module. The MFF module was therefore compared with the JPU method \cite{wu2019fastfcn}. From Table \ref{tab:5}, it can be seen that the proposed MFF module obtains better results than the JPU method\cite{wu2019fastfcn}.

\subsection{Ablation Study}
In order to verify the effectiveness of each module in the proposed difference-aware decoder based on Res2Net-50, ablation experiments were conducted on two COD datasets (CAMO \cite{le2019anabranch}, COD10K \cite{fan2020camouflaged}) and two SOD datasets (ECSSD\cite{yan2013hierarchical}, PASCAL-S\cite{li2014secrets}).

\subsubsection{Field Expansion Module (FEM)}
\begin{table}[htb]
\centering
\caption{Comparison of different receptive field methods in the FEM.}
\label{tab:3}
\small
\setlength\tabcolsep{1pt}
\begin{tabular}{c|cc|cc|cc|cc}
\hline
\multirow{2}{*}{Methods}
& \multicolumn{2}{c|}{CAMO} & \multicolumn{2}{c|}{COD10K} & \multicolumn{2}{c|}{ECSSD} & \multicolumn{2}{c}{PASCAL-S}
\\ \cline{2-9}
& $E_{\phi}\uparrow$ & $M\downarrow$
& $E_{\phi}\uparrow$ & $M\downarrow$
& $E_{\phi}\uparrow$ & $M\downarrow$
& $E_{\phi}\uparrow$ & $M\downarrow$  \\ \hline
RFB \cite{liu2018receptive}
& 0.875 & 0.073
& 0.901 & 0.033
& 0.957 & 0.030
& 0.904 & 0.059\\
ASPP \cite{chen2017deeplab}
& 0.888 & 0.068
& 0.905 & 0.032
& 0.957 & 0.029
& 0.909 & 0.057\\
D-ASPP\cite{yang2018denseaspp}
& 0.883 & 0.070
& 0.904 & 0.032
& 0.957 & 0.029
& 0.905 & 0.060 \\
FEM(w/o DR)
& 0.887 & 0.068
& \textbf{0.906} & \textbf{0.032}
& 0.960 & \textbf{0.028}
& 0.908 & 0.059 \\\hline
FEM(proposed)
& \textbf{0.895} & \textbf{0.063}
& 0.905 & \textbf{0.032}
& \textbf{0.963} & \textbf{0.028}
& \textbf{0.909} & \textbf{0.058} \\ \hline
\end{tabular}
\end{table}
The FEM was used in Stages A and B to enhance the receptive field of the feature map, so as to obtain sufficient contextual information to establish the dependency between the foreground and the background. As is well known, modules such as ASPP \cite{chen2017deeplab}, Dense ASPP\cite{yang2018denseaspp}, and RFB \cite{liu2018receptive} can also have the same effect. Therein, we replaced the proposed FEM with ASPP/FRB for the ablation study. From Table \ref{tab:3}, it can be seen that the proposed FEM can achieve the best accuracy in the COD/SOD tasks. Furthermore, we removed the dilation rate in the FEM (w/o DR) and also found that the performance dropped. It can be concluded that the consecutive atrous convolutions with different dilation rates are helpful, while the proposed FEM can outperform the other well-known modules designed for the enhancement of the receptive field.

\subsubsection{Difference Guidance Module (DGM)}
\begin{table}[htb]
\centering
\caption{Ablation study for the DAE (Stage C) with/without the DGM.}
\label{tab:6}
\small
\setlength\tabcolsep{1pt}
\begin{tabular}{c|cc|cc|cc|cc}
\hline
\multirow{2}{*}{Methods}
& \multicolumn{2}{c|}{CAMO} & \multicolumn{2}{c|}{COD10K} & \multicolumn{2}{c|}{ECSSD} & \multicolumn{2}{c}{PASCAL-S}
\\ \cline{2-9}
& $E_{\phi}\uparrow$ & $M\downarrow$
& $E_{\phi}\uparrow$ & $M\downarrow$
& $E_{\phi}\uparrow$ & $M\downarrow$
& $E_{\phi}\uparrow$ & $M\downarrow$  \\ \hline
w/o DGM
& 0.887 & 0.068
& 0.898 & 0.038
& 0.950 & 0.032
& 0.901 & 0.060
\\ \hline
DGM(proposed)
& \textbf{0.895} & \textbf{0.063}
& \textbf{0.905} & \textbf{0.032}
& \textbf{0.963} & \textbf{0.028}
& \textbf{0.909}  & \textbf{0.058} \\ \hline
\end{tabular}
\end{table}


In Stage C, the DGM is utilized to enhance the foreground features from the two branches. The DGM can be regarded as a plug-and-play module. Table \ref{tab:6} lists the results of the difference-aware decoder with/without the DGM. It can be seen that the features are enhanced by the guide map in a way similar to cross-attention, and the prior knowledge of the foreground and the background on the guide map can be used to complete the screening of the features, which is the key to improving the model's performance.

\subsubsection{Difference Enhancement Module (DEM)}
\begin{table}[htb]
\centering
\caption{Ablation study for the DEM (Satge C) with foreground (F), background (B), and the fusion of foreground and background (F-B).}
\label{tab:7}
\normalsize
\setlength\tabcolsep{1pt}
\begin{tabular}{c|cc|cc|cc|cc}
\hline
\multirow{2}{*}{Methods}
& \multicolumn{2}{c|}{CAMO} & \multicolumn{2}{c|}{COD10K} & \multicolumn{2}{c|}{ECSSD} & \multicolumn{2}{c}{PASCAL-S}
\\ \cline{2-9}
& $E_{\phi}\uparrow$ & $M\downarrow$
& $E_{\phi}\uparrow$ & $M\downarrow$
& $E_{\phi}\uparrow$ & $M\downarrow$
& $E_{\phi}\uparrow$ & $M\downarrow$  \\ \hline
F
& 0.868 & 0.073
& 0.903 & 0.033
& 0.958 & 0.029
& 0.905 & 0.061
\\
B
& 0.885 & 0.070
& 0.900 & 0.033
& 0.954 & 0.031
& \textbf{0.910} & \textbf{0.058}
\\ \hline
F-B(proposed)
& \textbf{0.895} & \textbf{0.063}
& \textbf{0.905} & \textbf{0.032}
& \textbf{0.963} & \textbf{0.028}
& \textbf{0.909} & \textbf{0.058} \\ \hline
\end{tabular}
\end{table}
Also in Stage C, the DEM is proposed to further enhance the difference between the foreground and the background. We chose the simple foreground ($F$), the background ($B$), and the subtraction of $F-B$ as the output features. From Table \ref{tab:7}, it is clear that $F-B$ can achieve the best results, which validates the original intent of the proposed fusion design. It can be concluded that the difference-aware fusion of the two can more efficiently distinguish the background and precisely extract the foreground objects.


\subsubsection{Repeat of DAE}
\begin{table}[htb]
\centering
\caption{Ablation study for the DAE with different repeat times.}
\label{tab:8}
\normalsize
\setlength\tabcolsep{2pt}
\begin{tabular}{c|cc|cc|cc|cc}
\hline
\multirow{2}{*}{Methods}
& \multicolumn{2}{c|}{CAMO} & \multicolumn{2}{c|}{COD10K} & \multicolumn{2}{c|}{ECSSD} & \multicolumn{2}{c}{PASCAL-S}
\\ \cline{2-9}
& $E_{\phi}\uparrow$ & $M\downarrow$
& $E_{\phi}\uparrow$ & $M\downarrow$
& $E_{\phi}\uparrow$ & $M\downarrow$
& $E_{\phi}\uparrow$ & $M\downarrow$  \\ \hline
1
& 0.887 & 0.068
& 0.902 & 0.033
& 0.955 & 0.029
& 0.907 & 0.058 \\
3
& 0.881 & 0.071
& 0.904 & \textbf{0.032}
& 0.955 & 0.030
& \textbf{0.909} & 0.059
\\ \hline
2(proposed)
& \textbf{0.895} & \textbf{0.063}
& \textbf{0.905} & \textbf{0.032}
& \textbf{0.963} & \textbf{0.028}
& \textbf{0.909} & \textbf{0.058} \\ \hline
\end{tabular}
\end{table}
The DAE used in Stage C can be regarded as a plug-and-play module, and can be utilized multiple times. As can be observed in Table \ref{tab:8}, repeating the DAE three times can lead to a significant performance degradation. From our understanding, the map generated from the output of the DAE is regarded as the guide map for the next extractor. However, the map from the output of the DAE is very similar to that of the feature map from the MFF module. Therein, the DAE fails to improve the difference features. From Table \ref{tab:8}, it is apparent that repeating the DAE twice obtains the best results.

\section{Conclusion}
In this paper, inspired by the way human eyes detect objects of interest, we have proposed a unified dual-branch decoder paradigm dubbed the difference-aware decoder. The proposed difference-aware decoder consists of three stages. Stage A is used to generate a guide map from layers 1 and 5 of the backbone. In Stage B, layers 2~4 are used to generate the background-aware features. In Stage C, the two features are fused to generate the enhanced feature maps between the foreground and background. The many experiments conducted in this study confirmed the superiority of the proposed difference-aware decoder for SOD, COD, polyp segmentation, and mirror detection tasks. Furthermore, we also proved the effectiveness of the different modules, including the FEM in Stage A, the MFF module in Stage B, and the DGM and DEM in Stage C. In the future, we will extend the proposed difference-aware decoder to other binary segmentation tasks, such as road extraction.


%





\ifCLASSOPTIONcaptionsoff
  \newpage
\fi



\bibliographystyle{IEEEtran}
\bibliography{IEEEabrv,ref}

\clearpage
\end{document}